\documentclass{article}[12pt]
\usepackage[margin=1in]{geometry}

\usepackage{amsmath,amsfonts,amsthm}
\usepackage{mathtools}
\usepackage{graphicx}        
\usepackage[bottom]{footmisc}
\usepackage{algorithm}
\usepackage{algorithmic}
\usepackage{url}

\makeindex             

\newcommand{\Normal}{\mathrm{Normal}}
\newcommand{\Cov}{\mathrm{Cov}}
\newcommand{\ybar}{\overline{y}}
\newcommand{\Var}{\mathrm{Var}}
\newcommand{\xpred}{x^*}
\newcommand{\EI}{\mathrm{EI}}
\newcommand{\KG}{\mathrm{KG}}
\newcommand{\argmax}{\mathop{\mathrm{argmax}}}

\newcommand{\optional}[1]{}
\newtheorem{proposition}{Proposition}

\begin{document}

\title{Bayesian optimization for materials design}
\author{Peter I. Frazier, Jialei Wang\\
pf98@cornell.edu, jw865@cornell.edu\\
School of Operations Research \& Information Engineering\\
Cornell University\\
Ithaca, NY 14853
}
\date{}
\maketitle

\abstract*{
We introduce Bayesian optimization, a technique developed for optimizing time-consuming engineering simulations and for fitting machine learning models on large datasets.  Bayesian optimization guides the choice of experiments during materials design and discovery to find good material designs in as few experiments as possible.  We focus on the case when materials designs are parameterized by a low-dimensional vector. Bayesian optimization is built on a statistical technique called Gaussian process regression, which allows predicting the performance of a new design based on previously tested designs.  After providing a detailed introduction to Gaussian process regression, we introduce two Bayesian optimization methods: expected improvement, for design problems with noise-free evaluations; and the knowledge-gradient method, which generalizes expected improvement and may be used in design problems with noisy evaluations. Both methods are derived using a value-of-information analysis, and enjoy one-step Bayes-optimality.
}


\section{Introduction}
In materials design and discovery, we face the problem of choosing the chemical structure, composition, or processing conditions of a material to meet design criteria.
The traditional approach is to use iterative trial and error, in which we (1) choose some material design that we think will work well based on intuition, past experience, or theoretical knowledge; (2) synthesize and test the material in physical experiments; and (3) use what we learn from these experiments in choosing the material design to try next.  This iterative process is repeated until some combination of success and exhaustion is achieved.

While trial and error has been extremely successful, we believe that mathematics and computation together promise to accelerate the pace of materials discovery, not by changing the fundamental iterative nature of materials design, but by improving the choices that we make about which material designs to test, and by improving our ability to learn from previous experimental results.

In this chapter, we describe a collection of mathematical techniques, based on Bayesian statistics and decision theory, for augmenting and enhancing the trial and error process.
We focus on one class of techniques, called Bayesian optimization (BO), or Bayesian global optimization (BGO), which use machine learning to build a predictive model of the underlying relationship between the design parameters of a material and its properties, and then use decision theory to suggest which design or designs would be most valuable to try next. The most well-developed Bayesian optimization methods assume that (1) the material is described by a vector of continuous variables, as is the case, e.g., when choosing ratios of constituent compounds, or choosing a combination of temperature and pressure to use during manufacture; (2) we have a single measure of quality that we wish to make as large as possible; and (3) the constraints on feasible materials designs are all , so that any unknown constraints are incorporated into the quality measure.  There is also a smaller body of work on problems that go beyond these assumptions, either by considering discrete design decision (such as small molecule design), multiple competing objectives, or by explicitly allowing unknown constraints.

Bayesian optimization was pioneered by \cite{Ku64}, with early development through the 1970s and 1980s by Mockus and Zilinskas \cite{MoTiZi78,Mo89}.  Development in the 1990s was marked by the popularization of Bayesian optimization by Jones, Schonlau, and Welch, who, building on previous work by Mockus, introduced the Efficient Global Optimization (EGO) method \cite{JoScWe98}.  This method became quite popular and well-known in engineering, where it has been adopted for design applications involving time-consuming computer experiments, within a broader set of methods designed for optimization of expensive functions \cite{BookerDennis1998}.  In the 2000s, development of Bayesian optimization continued in statistics and engineering, and the 2010s have seen additional development from the machine learning community, where Bayesian optimization is used for tuning hyperparameters of computationally expensive machine learning models \cite{snoek2012practical}.
Other introductions to Bayesian optimization may be found in the tutorial article \cite{BrCoFr09} and textbooks \cite{FoSoKe08,SaWiNo03}, and an overview of the history of the field may be found in \cite{Sa02}.

We begin in Section~\ref{sec:BO} by introducing the precise problem considered by Bayesian Optimization. We then describe in Section~\ref{sec:GP} the predictive technique used by Bayesian Optimization, which is called Gaussian Process (GP) regression.
We then show, in Section~\ref{sec:recommendation}, how Bayesian Optimization recommends which experiments to perform.
In Section~\ref{sec:software} we provide an overview of software packages, both freely available and commercial, that implement the Bayesian Optimization methods described in this chapter.
We offer closing remarks in Section~\ref{sec:conclusion}.

\section{Bayesian Optimization}
\label{sec:BO}

Bayesian optimization considers materials designs parameterized by a $d$-dimensional vector $x$.  We suppose that the space of materials designs in which $x$ takes values is a known set $A \subseteq \mathbb{R}^d$.

For example, $x=(x(1),\ldots,x(d))$ could give the ratio of each of $d$ different constituents mixed together to create some aggregate material. In this case, we would choose $A$ to be the set $A=\{x : \sum_{i=1}^d x(i) = 1 \}$.
As another example, setting $d=2$, $x=(x(1),x(2))$ could give the temperature ($x(1)$) and pressure ($x(2)$) used in material processing.
In this case, we would choose $A$ to be the rectangle bounded by the experimental setup's minimum and maximal achievable temperature on one axis, $T_{\mathrm{min}}$ and $T_{\mathrm{max}}$, and the minimum and maximum achievable pressure on the other.
As a final example, we could let $x=(x(1),\ldots,x(d)$ be the temperatures used in some annealing schedule, assumed to be decreasing over time.  In this case, we would set $A$ to be the set $\left\{x : T_{\mathrm{max}} \ge x(1) \ge \cdots \ge x(d) \ge T_{\mathrm{min}} \right\}$.

Let $f(x)$ be the quality of the material with design parameter $x$.  The function $f$ is unknown, and observing $f(x)$ requires synthesizing material design $x$ and observing its quality in a physical experiment.
We would like to find a design $x$ for which $f(x)$ is large.  That is, we would like to solve
\begin{equation}
\label{eq:obj}
\max_{x\in A} f(x).
\end{equation}
This is challenging because evaluating $f(x)$ is typically expensive and time-consuming.  While the time and expense depends on the setting, synthesizing and testing a new material design could easily take days or weeks of effort and thousands of dollars of materials.

In Bayesian optimization, we use mathematics to build a predictive model for the function $f$ based on observations of previous materials designs, and then use this predictive model to recommend a materials design that would be most valuable to test next.
We first describe this predictive model in Section~\ref{sec:GP}, which is performed using a machine learning technique called Gaussian process regression.
We then describe, in Section~\ref{sec:recommendation}, how this predictive model is used to recommend which design to test next.

\section{Gaussian Process regression}
\label{sec:GP}

The predictive piece of Bayesian optimization is based on a machine learning technique called Gaussian process regression.
This technique is a Bayesian version of a frequentist technique called kriging, introduced in the geostatistics literature by South-African mining engineer Daniel Krige \cite{kbiob1951statistical}, and popularized later by Matheron and colleagues \cite{matheron1971theory}, as described in \cite{cressie1990origins}. A modern monograph on Gaussian process regression is \cite{RaWi06}, and a list of software implementing Gaussian process regression may be found at \cite{GPwebsite}.

In Gaussian process regression, we seek to predict $f(x)$ based on observations at previously evaluated points, call them $x_1,\ldots,x_n$.
We first treat the case where $f(x)$ can be observed exactly, without noise, and then later treat noise in Section~\ref{sec:GP-noise}.
In this noise-free case, our observations are $y_i = f(x_i)$ for $i=1,\ldots,n$.

Gaussian process regression is a Bayesian statistical method, and in Bayesian statistics we perform inference by placing a so-called {\it prior probability distribution} on unknown quantities of interest.  The prior probability distribution is often called, more simply, the {\it prior distribution} or, even more simply, the {\it prior}.  This prior distribution is meant to encode our intuition or domain expertise regarding which values for the unknown quantity of interest are most likely.  We then use Bayes rule, together with any data observed, to calculate a {\it posterior probability distribution} on these unknowns.  For a broader introduction to Bayesian statistics, see the textbook \cite{GeCaStRu04} or the research monograph \cite{Be85}.

In Gaussian process regression, if we wish to predict the value of $f$ at a single candidate point $\xpred$, it is sufficient to consider our unknowns to be the values of $f$ at the previously evaluated points, $x_1,\ldots,x_n$, and the new point $x^*$ at which we wish to predict.  That is, we take our unknown quantity of interest to be the vector $(f(x_1),\ldots,f(x_n),f(\xpred))$.  We then take our data, which is $f(x_1),\ldots,f(x_n)$, and use Bayes rule to calculate a posterior probability distribution on the full vector of interest, $(f(x_1),\ldots,f(x_n),f(\xpred))$, or, more simply, just on $f(\xpred)$.

To calculate the posterior, we must first specify the prior, which
Gaussian process regression assumes to be multivariate normal.  It calculates the mean vector of this multivariate normal prior distribution using a function, called the {\it mean function} and written here as $\mu_0(\cdot)$, which takes a single $x$ as an argument.  It applies this mean function to each of the points $x_1,\ldots,x_n,\xpred$ to create an $n+1$-dimensional column vector. 
Gaussian process regression creates the covariance matrix of the multivariate normal prior distribution using another function, called the {\it covariance function} or {\it covariance kernel} and written here as $\Sigma_0(\cdot,\cdot)$, which takes a pair of points $x,x'$ as arguments.
It applies this covariance function to every pair of points in $x_1,\ldots,x_n,x$ to create an $(n+1)\times (n+1)$ matrix. 

Thus, Gaussian process regression sets the prior probability distribution to,
\begin{equation}
\begin{bmatrix}
f(x_1) \\ \ldots \\ f(x_n) \\ f(\xpred)
\end{bmatrix}
\sim \Normal\left(
\begin{bmatrix}
\mu_0(x_1) \\ \ldots \\ \mu_0(x_n) \\ \mu_0(\xpred)
\end{bmatrix},
\begin{bmatrix}
\Sigma_0(x_1,x_1)    & \cdots & \Sigma_0(x_1,x_n)    & \Sigma_0(x_1,\xpred) \\
\vdots               & \ddots & \vdots               & \vdots \\
\Sigma_0(x_n,x_1)    & \cdots & \Sigma_0(x_n,x_n)    & \Sigma_0(x_n,\xpred) \\
\Sigma_0(\xpred,x_1) & \cdots & \Sigma_0(\xpred,x_n) & \Sigma_0(\xpred,\xpred)
\end{bmatrix}\right)
\label{eq:GP-prior-expanded}
\end{equation}
The subscript ``0'' in $\mu_0$ and $\Sigma_0$ indicate that these functions are relevant to the prior distribution, before any data has been collected.

We now discuss how the mean and covariance functions are chosen, focusing on the covariance function first because it tends to be more important in getting good results from Gaussian process regression.

\subsection{Choice of covariance function}
In choosing the covariance function $\Sigma_0(\cdot,\cdot)$, we wish to satisfy two requirements.

The first is that it should encode the belief that points $x$ and $x'$ near each other tend to have more similar values for $f(x)$ and $f(x')$.  To accomplish this, we want the covariance matrix in \eqref{eq:GP-prior-expanded} to have entries that are larger for pairs of points that are closer together, and closer to 0 for pairs of points that are further apart.

The second is that the covariance function should always produce positive semidefinite covariance matrices in the multivariate normal prior.
That is, if $\Sigma$ is the covariance matrix in \eqref{eq:GP-prior-expanded}, then we require that $a^T \Sigma a \ge 0$ for all column vectors $a$ (where $a$ is assumed to have the appropriate length, $n+1$). This requirement is necessary to ensure that the multivariate normal prior distribution is a well-defined probability distribution, because if $\theta$ is multivariate normal with mean vector $\mu$ and covariance matrix $\Sigma$, then the variance of $a\cdot \theta$ is $a^T \Sigma a$, and we require variances to be non-negative.

Several covariance functions satisfy these two requirements.  The most commonly used is called the {\it squared exponential}, or Gaussian kernel, and is given by,
\begin{equation}
\Sigma_0(x,x') = \alpha \exp\left(-\sum_{i=1}^d \beta_i (x_i-x'_i)^2\right).
\label{eq:kernel}
\end{equation}
This kernel is parameterized by $d+1$ parameters: $\alpha$, and $\beta_1,\ldots,\beta_d$.

The parameter $\alpha>0$ controls how much overall variability there is in the function $f$.
We observe that under the prior, the variance of $f(x)$ is $\Var(f(x)) = \Cov(f(x),f(x)) = \alpha$.
Thus, when $\alpha$ is large, we are encoding in our prior distribution that $f(x)$ is likely to take a larger range of values.

The parameters $\beta_i > 0$ controls how quickly the function $f$ varies with $x$.
For example, consider the relationship between some point $x$ and another point $x'=x+[1,0,\ldots,0]$.
When $\beta_1$ is small (close to $0$), the covariance between $f(x)$ and $f(x')$ is 
$\alpha \exp(-\beta_1) \approx \alpha$, giving a correlation between $f(x)$ and $f(x')$
of nearly $1$.  This reflects a belief that $f(x)$ and $f(x')$ are likely to be very similar,
and that learning the value of $f(x)$ will also teach us a great deal about $f(x')$.
In contrast, when $\beta_1$ is large, the covariance between $f(x)$ and $f(x')$ is nearly $0$, given a correlation between $f(x)$ and $f(x')$ that is also nearly $0$, reflecting a belief that $f(x)$ and $f(x')$ are unrelated to each other, and learning something about $f(x)$ will teach us little about $(x')$.

\paragraph{Going beyond the squared exponential kernel}

There are several other possibilities for the covariance kernel beyond the squared exponential kernel, which encode different assumptions about the underlying behavior of the function $f$. One particularly useful generalization of the squared exponential covariance kernel is the Mat\'{e}rn covariance kernel, which allows more flexibility in modeling the smoothness of $f$.

Before describing this kernel, let $r = \sqrt{\sum_i \left(\frac{x_i - x'_i}{\beta_i}\right)^2}$ be the Euclidean distance between $x$ and $x'$, but where we have altered the length scale in each dimension by some strictly positive parameter $\beta_i$.  Then, the squared exponential covariance kernel can be written as, $\Sigma_0(x,x') = \alpha \exp\left(-r^2\right)$.

With this notation, the Mat\'{e}rn covariance kernel is,
\begin{equation*}
\Sigma_0(x,x') = \alpha \frac{2^{1-\nu}}{\Gamma(\nu)}\left(\sqrt{2\nu}r\right)^\nu K_\nu\left(\sqrt{2\nu} r\right),
\end{equation*}
where $K_\nu$ is the modified Bessel function.
If we take the limit as $\nu\to\infty$, we obtain the squared exponential kernel (\cite{RaWi06}, Section 4.2 page 85).

The Mat\'{e}rn covariance kernel is useful because it allows modeling the smoothness of $f$ in a more flexible way, as compared with the squared exponential kernel.
Under the squared exponential covariance kernel, the function $f$ is infinitely mean-square differentiable\footnote{Being ``mean-square differentiable'' at $x$ in the direction given by the unit vector $e_i$ means that the limit $\lim_{\delta\to 0} (f(x+\delta e_i) - f(x))/\delta$ exists in mean square.  Being ``$k$-times mean-square differentiable'' is defined analogously.}, which may not be an appropriate assumption in many applications.
In contrast, under the Mat\'{e}rn covariance kernel, $f$ is $k$-times mean-square differentiable if and only if $\nu>k$.  Thus, we can model a function that is twice differentiable but no more by choosing $\nu=5/2$, and a function that is once differentiable but no more by choosing $\nu=3/2$.


While the squared exponential and Mat\'{e}rn covariance kernels allow modeling a wide range of behaviors, and together represent a toolkit that will handle a wide variety of applications, there are other covariance kernels.  For a thorough discussion of these, see Chapter~4 of \cite{RaWi06}.


Both the Mat\'{e}rn and squared exponential covariance kernel require choosing parameters.  While it certainly is possible for one to choose the parameters $\alpha$ and $\beta_i$ (and $\nu$ in the case of Mat\'{e}rn) based on one's intuition about $f$, and what kinds of variability $f$ is likely to have in a particular application, it is more common to choose these parameters (especially $\alpha$ and $\beta_i$) adaptively, so as to best fit previously observed points.  We discuss this more below in Section~\ref{sec:parameter}. First, however, we discuss the choice of the mean function.

\subsection{Choice of mean function}
We now discuss choosing the mean function $\mu_0(\cdot)$.  Perhaps the most common choice is to simply set the mean function equal to a constant, $\mu$.  This constant must be estimated, along with parameters of the covariance kernel such as $\alpha$ and $\beta_i$, and is discussed in Section~\ref{sec:parameter}.

Beyond this simple choice, if one believes that there will be trends in $f$ that can be described in a parametric way, then it is useful to include trend terms into the mean function.  This is accomplished by choosing
\begin{equation*}
\mu_0(x) = \mu + \sum_{j=1}^J \gamma_j \Psi_j(x),
\end{equation*}
where $\Psi_j(\cdot)$ are known functions, and $\gamma_j \in \mathbb{R}$, along with $\mu\in\mathbb{R}$, are parameters that must be estimated.

A common choice for the $\Psi_j$, if one chooses to include them, are polynomials in $x$ up to some small order.
For example, if $d=2$, so $x$ is two-dimensional, then one might include all polynomials up to second order,
$\Psi_1(x) = x_1$, 
$\Psi_2(x) = x_2$, 
$\Psi_3(x) = (x_1)^2$,
$\Psi_4(x) = (x_2)^2$,
$\Psi_5(x) = x_1 x_2$, setting $J=5$.
One recovers the constant mean function by setting $J=0$.

\subsection{Inference}
\label{sec:inference}
Given the prior distribution \eqref{eq:GP-prior-expanded} on $f(x_1),\ldots,f(x_n),f(\xpred)$, and given (noise-free) observations of $f(x_1),$\dots, $f(x_n)$, the critical step in Gaussian process regression is calculating the posterior distribution on $f(\xpred)$.  
We rely on the following general result about conditional probabilities and multivariate normal distributions.  Its proof, which may be found in Section~\ref{sec:derivations}, relies on Bayes rule and algebraic manipulation of the probability density of the multivariate normal distribution.

\begin{proposition}
Let $\theta$ be a $k$-dimensional multivariate normal random column vector, with mean vector $\mu$ and covariance matrix $\Sigma$.
Let $k_1\ge1, k_2\ge1$ be two integers summing to $k$.  Decompose $\theta$, $\mu$ and $\Sigma$ as
\begin{equation*}
\theta = \begin{bmatrix}\theta_{[1]} \\ \theta_{[2]}\end{bmatrix}, \qquad
\mu = \begin{bmatrix}\mu_{[1]} \\ \mu_{[2]}\end{bmatrix},
\qquad
\Sigma = \begin{bmatrix}\Sigma_{[1,1]} & \Sigma_{[1,2]} \\ \Sigma_{[2,1]} & \Sigma_{[2,2]}\end{bmatrix},
\end{equation*}
so that 
$\theta_{[i]}$ and $\mu_{[i]}$ are $k_i$-column vectors, and 
$\Sigma_{[i,j]}$ is a $k_i\times k_j$ matrix, for each $i,j=1,2$.

If $\Sigma_{1,1}$ and $\Sigma_{2,2}$ are invertible, then, for any $u\in \mathbb{R}^{k_1}$, the conditional distribution of $\theta_{[2]}$ given that $\theta_{[1]}=u$ is multivariate normal with mean 
\begin{equation*}
\mu_{[2]} + \Sigma_{[2,1]}\Sigma_{[1,1]}^{-1}(u-\mu_{[1]})
\end{equation*}
and covariance matrix
\begin{equation*}
\Sigma_{[2,2]} - \Sigma_{[2,1]}\Sigma_{[1,1]}^{-1}\Sigma_{[1,2]}.
\end{equation*}
\label{prop:multivariate-normal-conditioning}
\end{proposition}

We use this proposition to calculate the posterior distribution on $f(\xpred)$, given $f(x_1),\ldots,f(x_n)$.

Before doing so, however, we first introduce some additional notation. We let $y_{1:n}$ indicate the column vector $[y_1,\ldots,y_n]^T$, and we let $x_{1:n}$ indicate the sequence of vectors $(x_1,\ldots,x_n)$.
We let $f(x_{1:n}) = [f(x_1),\ldots,f(x_n)]^T$, and similarly for other functions of $x$, such as $\mu_0(\cdot)$.  We introduce similar additional notation for functions that take pairs of points $x,x'$, so that 
$\Sigma(x_{1:n},x_{1:n})$ is the matrix
\begin{equation*}
\left[
\begin{smallmatrix}
\Sigma_0(x_1,x_1) & \cdots & \Sigma_0(x_1,x_n) \\
\vdots & \ddots & \vdots \\
\Sigma_0(x_n,x_1) & \cdots & \Sigma_0(x_n,x_n)
\end{smallmatrix}
\right],
\end{equation*}
$\Sigma_0(\xpred,x_{1:n})$ is the row vector 
$[\Sigma_0(\xpred,x_1),\ldots,\Sigma_0(\xpred,x_n)]$,
and 
$\Sigma_0(x_{1:n},\xpred)$ is the column vector \\
$[\Sigma_0(x_1,\xpred),\ldots,\Sigma_0(x_n,\xpred)]^T$.

This notation allows us to rewrite \eqref{eq:GP-prior-expanded} as
\begin{equation}
\label{eq:GP-prior}
\begin{bmatrix}
y_{1:n}\\ f(\xpred)
\end{bmatrix}
=
\Normal\left(
\begin{bmatrix}
\mu_0(x_{1:n})\\ \mu_0(\xpred)
\end{bmatrix},
\begin{bmatrix}
\Sigma_0(x_{1:n},x_{1:n}) & \Sigma_0(x_{1:n},\xpred) \\
\Sigma_0(\xpred,x_{1:n})  & \Sigma_0(\xpred,\xpred)
\end{bmatrix}
\right).
\end{equation}

We now examine this expression in the context of Proposition~\ref{prop:multivariate-normal-conditioning}. We set $\theta_{[1]} = f(x_{1:n})$, $\theta_{[2]} = f(\xpred)$,
$\mu_{[1]} = \mu_0(x_{1:n})$, 
$\mu_{[2]} = \mu_0(\xpred)$, 
$\Sigma_{[1,1]} = \Sigma_0(x_{1:n},x_{1:n})$,
$\Sigma_{[1,2]} = \Sigma_0(x_{1:n},\xpred)$,
$\Sigma_{[2,1]} = \Sigma_0(\xpred,x_{1:n})$, and
$\Sigma_{[2,2]} = \Sigma_0(\xpred,\xpred)$.

Then, applying Proposition~\ref{prop:multivariate-normal-conditioning}, we see that
the posterior distribution on $f(\xpred)$ given observations $y_i = f(x_i), i=1,\ldots,n$ is normal, with a mean $\mu_n(\xpred)$ and variance $\sigma^2_n(\xpred)$ given by,
\begin{align}
\mu_n(\xpred) &= \mu_0(\xpred) + \Sigma_0(\xpred,x_{1:n})\Sigma_0(x_{1:n},x_{1:n})^{-1}(f(x_{1:n})-\mu_0(x_{1:n})),
\label{eq:GP-noise-free-posterior-mean} \\
\sigma^2_n(\xpred) &= \Sigma_0(\xpred,\xpred) - \Sigma_0(\xpred,x_{1:n})\Sigma_0(x_{1:n},x_{1:n})^{-1}\Sigma_0(x_{1:n},\xpred).
\label{eq:GP-noise-free-posterior-covariance}
\end{align}

The invertibility of $\Sigma_0(x_{1:n},x_{1:n})$ (and also $\Sigma_0(x^*,x^*)$) required by Proposition~\ref{prop:multivariate-normal-conditioning} depends on the covariance kernel and its parameters (typically called hyperparameters), but this invertibility typically holds as long as these hyperparameters satisfy mild non-degeneracy conditions, and the $x_{1:n}$ are distinct, i.e., that we have not measured the same point more than once.  For example, under the squared exponential covariance kernel, invertibility holds as long as $\alpha>0$ and the $x_{1:n}$ are distinct. If we have measured a point multiple times, then we can safely drop all but one of the measurements, here where observations are noise-free. Below, we treat the case where observations are noisy, and in this case including multiple measurements of the same point is perfectly reasonable and does not cause issues.

\begin{figure}
    \begin{center}
\includegraphics[scale=0.41]{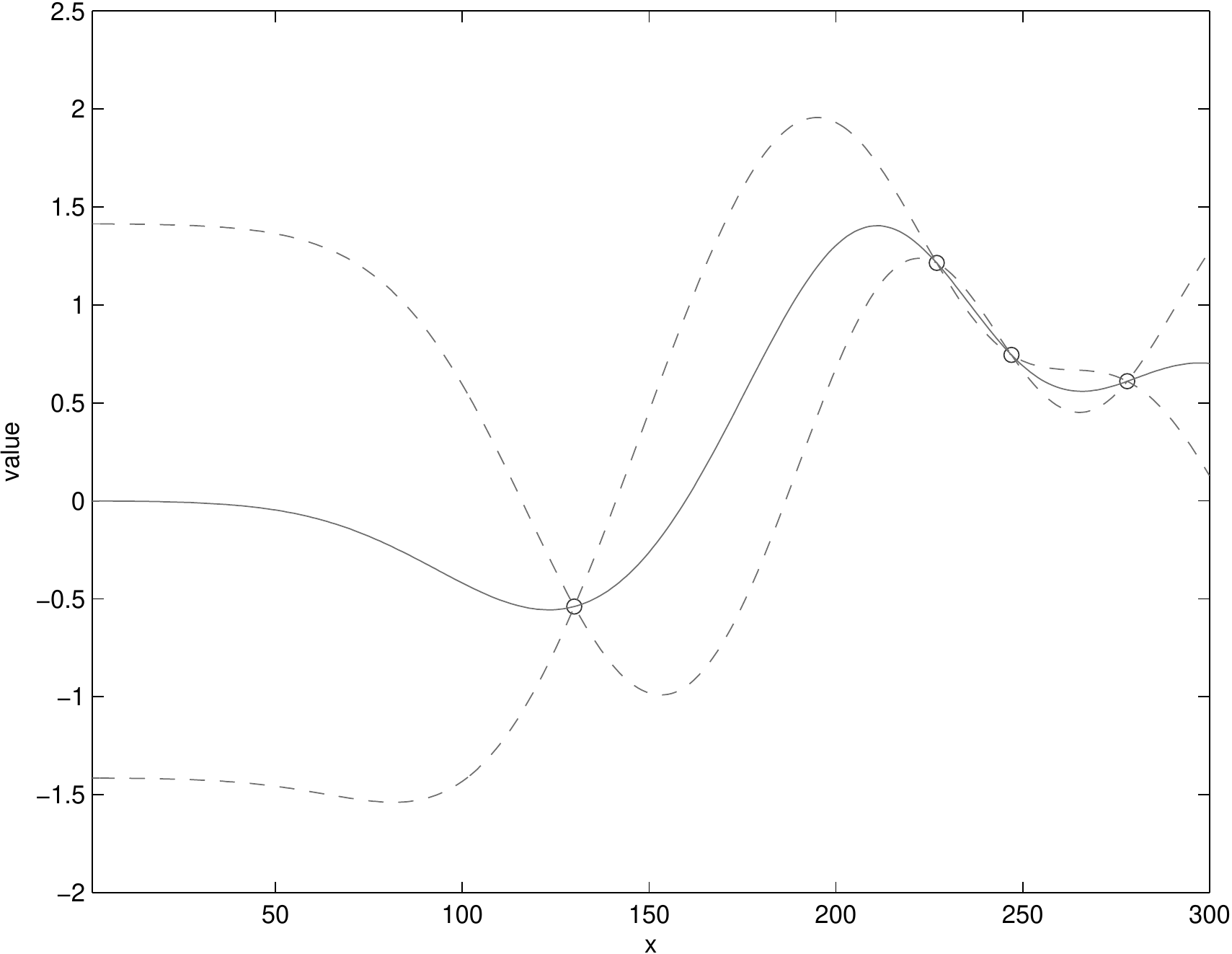}
    \end{center}
\caption{Illustration of Gaussian process regression with noise-free evaluations.  The circles show previously evaluated points, $(x_i,f(x_i))$.  The solid line shows the posterior mean, $\mu_n(x)$ as a function of $x$, which is an estimate $f(x)$, and the dashed lines show a Bayesian credible interval for each $f(x)$, calculated as $\mu_n(x) \pm 1.96\sigma_n(x)$.
Although this example shows $f$ taking a scalar input, Gaussian process regression can be used for functions with vector inputs.
\label{fig:GP-noise-free}}
\end{figure}

Figure~\ref{fig:GP-noise-free} shows the output from Gaussian process regression.  In the figure, circles show points $(x_i,f(x_i))$, the solid line shows $\mu_n(\xpred)$ as a function of $\xpred$, and the dashed lines are positioned at $\mu_n(\xpred) \pm 1.96\sigma_n(\xpred)$, forming a 95\% Bayesian credible interval for $f(\xpred)$, i.e., an interval in which $f(\xpred)$ lies with posterior probability 95\%.  (A credible interval is the Bayesian version of a frequentist confidence interval.)
Because observations are noise-free, the posterior mean $\mu_n(\xpred)$ interpolates the observations $f(\xpred)$.

\subsection{Inference with just one observation}
\label{sec:just-one}

The expressions \eqref{eq:GP-noise-free-posterior-mean} and \eqref{eq:GP-noise-free-posterior-covariance} are complex, and perhaps initially difficult to assimilate.  To give more intuition about them, and also to support some additional analysis below in Section~\ref{sec:recommendation}, it is useful to consider the simplest case, when we have just a single measurement, $n=1$.

In this case, all matrices in \eqref{eq:GP-noise-free-posterior-mean} and \eqref{eq:GP-noise-free-posterior-covariance} are scalars, $\Sigma_0(\xpred,x_1) = \Sigma_0(x_1,\xpred)$, and the expressions \eqref{eq:GP-noise-free-posterior-mean} and \eqref{eq:GP-noise-free-posterior-covariance} can be rewritten as,
\begin{align}
\mu_1(\xpred) &= \mu_0(\xpred) + \frac{\Sigma_0(\xpred,x_1)}{\Sigma_0(x_1,x_1)}(f(x_1)-\mu_0(x_1)), \label{eq:GP-intuition-mean}\\
\sigma^2_1(\xpred) &= \Sigma_0(\xpred,\xpred) - \frac{\Sigma_0(\xpred,x_1)^2}{\Sigma_0(x_1,x_1)}. \label{eq:GP-intuition-variance}
\end{align}

\paragraph{Intuition about the expression for the posterior mean}
We first examine \eqref{eq:GP-intuition-mean}.
We see that the posterior mean of $f(\xpred)$, $\mu_1(\xpred)$, which we can think of as our estimate of $f(\xpred)$ after observing $f(x_1)$, is obtained by taking our original estimate of $f(\xpred)$, $\mu_0(\xpred)$, and adding to it a correction term.  This correction term is itself the product of two quantities:  the error $f(x_1)-\mu_0(x_1)$ in our original estimate of $f(x_1)$, and the quantity $\frac{\Sigma_0(\xpred,x_1)}{\Sigma_0(x_1,x_1)}$.  Typically, $\Sigma_0(\xpred,x_1)$ will be positive, and hence also $\frac{\Sigma_0(\xpred,x_1)}{\Sigma_0(x_1,x_1)}$. 
(Recall, $\Sigma_0(x_1,x_1)$ is a variance, so is never negative.) Thus, if $f(x_1)$ is bigger than expected, $f(x_1) - \mu_0(x_1)$ will be positive, and our posterior mean $\mu_1(\xpred)$ will be larger than our prior mean $\mu_0(\xpred)$. In contrast, if $f(x_1)$ is smaller than expected, $f(x_1) - \mu_0(x_1)$ will be negative, and our posterior mean $\mu_1(\xpred)$ will be smaller than our prior mean $\mu_0(\xpred)$.

We can examine the quantity $\frac{\Sigma_0(\xpred,x_1)}{\Sigma_0(x_1,x_1)}$ to understand the effect of the position of $\xpred$ relative to $x_1$ on the magnitude of the correction to the posterior mean.  Notice that $\xpred$ only enters this expression through the numerator. If $\xpred$ is close to $x_1$, then $\Sigma_0(\xpred,x_1)$ will be large under the squared exponential and most other covariance kernels, and positive values for $f(x_1)-\mu_0(x_1)$ will also cause a strong positive change in $\mu_1(\xpred)$ relative to $\mu_0(\xpred)$. If $\xpred$ is far from $x_1$, then $\Sigma_0(\xpred,x_1)$ will be close to $0$, and $f(x_1)-\mu_0(x_1)$ will have little effect on $\mu_1(\xpred)$.

\paragraph{Intuition about the expression for the posterior variance}
Now we examine \eqref{eq:GP-intuition-variance}.
We see that the variance of our belief on $f(\xpred)$ under the posterior, $\sigma^2_1(\xpred)$, is smaller than its value under the prior, $\Sigma_0(\xpred,\xpred)$.  Moreover, when $\xpred$ is close to $x_1$, $\Sigma_0(\xpred,x_1)$ will be large, and the reduction in variance from prior to posterior will also be large. 

Conversely, when $\xpred$ is far from $x_1$, $\Sigma_0(\xpred,x_1)$ will be close to $0$, and the variance under the posterior will be similar to its value under the prior.

As a final remark, we can also rewrite the expression \eqref{eq:GP-intuition-variance} in terms of the squared correlation under the prior, 
$\mathrm{Corr}(f(\xpred),f(x_1))^2 = \Sigma_0(\xpred,x_1)/(\Sigma_0(\xpred,\xpred)\Sigma_0(x_1,x_1)) \in [0,1]$, as 
\begin{equation*}
\sigma^2_1(\xpred) = \Sigma_0(\xpred,\xpred)\left( 1 - \mathrm{Corr}(f(\xpred),f(x_1))^2\right).
\end{equation*}
We thus see that the reduction in variance of the posterior distribution depends on the squared correlation under the prior, with larger squared correlation implying a larger reduction.

\subsection{Inference with noisy observations}
\label{sec:GP-noise}
The previous section assumed that $f(\xpred)$ can be observed exactly, without any error.
When $f(\xpred)$ is the outcome of a physical experiment, however,
our observations are obscured by noise.  Indeed, if we were to synthesize and test the same material design $\xpred$ multiple times, we might observe different results.

To model this situation, Gaussian process regression can be extended to allow observations of the form,
\begin{equation*}
y(x_i) = f(x_i) + \epsilon_i, \label{eq:noise}
\end{equation*}
where we assume that the $\epsilon_i$ are normally distributed with mean $0$ and constant variance, $\lambda^2$, with independence across $i$.
In general, the variance $\lambda^2$ is unknown, but we treat it as a known parameter of our model, and then estimate it along with all the other parameters of our model, as discussed below in Section~\ref{sec:parameter}.

These assumptions of constant variance (called {\it homoscedasticity}) and independence make the analysis significantly easier, although they are often violated in practice.  Experimental conditions that tend to violate these assumptions are discussed below, as are versions of GP regression that can be used when they are violated.

\paragraph{Analysis of independent homoscedastic noise}

To performance inference under independent homoscedastic noise, and calculate a posterior distribution on the value of the function $f(x_*)$ at a given point $x_*$, our first step is to write down the joint distribution of our observations $y_1,\ldots,y_n$ and the quantity we wish to predict, $f(x_*)$, under the prior.
That is, we write down the distribution of the vector $[y_1,\ldots,y_n,f(x_*)]$.

We first observe that $[y_1,\ldots,y_n,f(x_*)]$ is the sum of $[f(x_1),\ldots,f(x_n),f(x_*)]$ and another vector, $[\epsilon_1,\ldots,\epsilon_n,0]$.  The first vector has a multivariate normal distribution given by \eqref{eq:GP-prior}.  The second vector is independent of the first and is also multivariate normal, with a mean vector that is identically $0$, and a covariance matrix $\mathrm{diag}(\lambda^2,\ldots,\lambda^2,0)$.
The sum of two independent multivariate normal random vectors is itself multivariate normal, with a mean vector and covariance matrix given, respectively, by the sums of the mean vectors and covariance matrices of the summands.  This gives the distribution of $[y_1,\ldots,y_n,f(x_*)]$ as
\begin{equation}
\begin{bmatrix}
y_{1:n}\\ f(\xpred)
\end{bmatrix}
\sim
\Normal\left(
\begin{bmatrix}
\mu_0(x_{1:n})\\ \mu_0(\xpred)
\end{bmatrix},
\begin{bmatrix}
\Sigma_0(x_{1:n},x_{1:n})+\lambda^2 I_n & \Sigma_0(x_{1:n},\xpred) \\
\Sigma_0(\xpred,x_{1:n})                & \Sigma_0(\xpred,\xpred)
\end{bmatrix}
\right),
\label{eq:GP-prior-noise}
\end{equation}
where $I_n$ is the $n$-dimensional identity matrix.

As we did in Section~\ref{sec:inference}, we can use Proposition~\ref{prop:multivariate-normal-conditioning} with the above expression to compute the posterior on $f(\xpred)$ given $f(x_{1:n})$.  We obtain,
\begin{align}
\mu_n(\xpred) &= \mu_0(\xpred) + \Sigma_0(\xpred,x_{1:n})\left[\Sigma_0(x_{1:n},x_{1:n})+\lambda^2 I_n\right]^{-1}(y_{1:n}-\mu_0(x_{1:n}))
\label{eq:GP-noisy-posterior-mean} \\
\sigma^2_n(\xpred) &= \Sigma_0(\xpred,\xpred) - \Sigma_0(\xpred,x_{1:n})\left[\Sigma_0(x_{1:n},x_{1:n})+\lambda^2 I_n\right]^{-1}\Sigma_0(x_{1:n},\xpred).
\label{eq:GP-noisy-posterior-covariance}
\end{align}

If we set $\lambda^2=0$, so there is no noise, then we recover \eqref{eq:GP-noise-free-posterior-mean} and \eqref{eq:GP-noise-free-posterior-covariance}.

\begin{figure}
    \begin{center}
\includegraphics[scale=0.41]{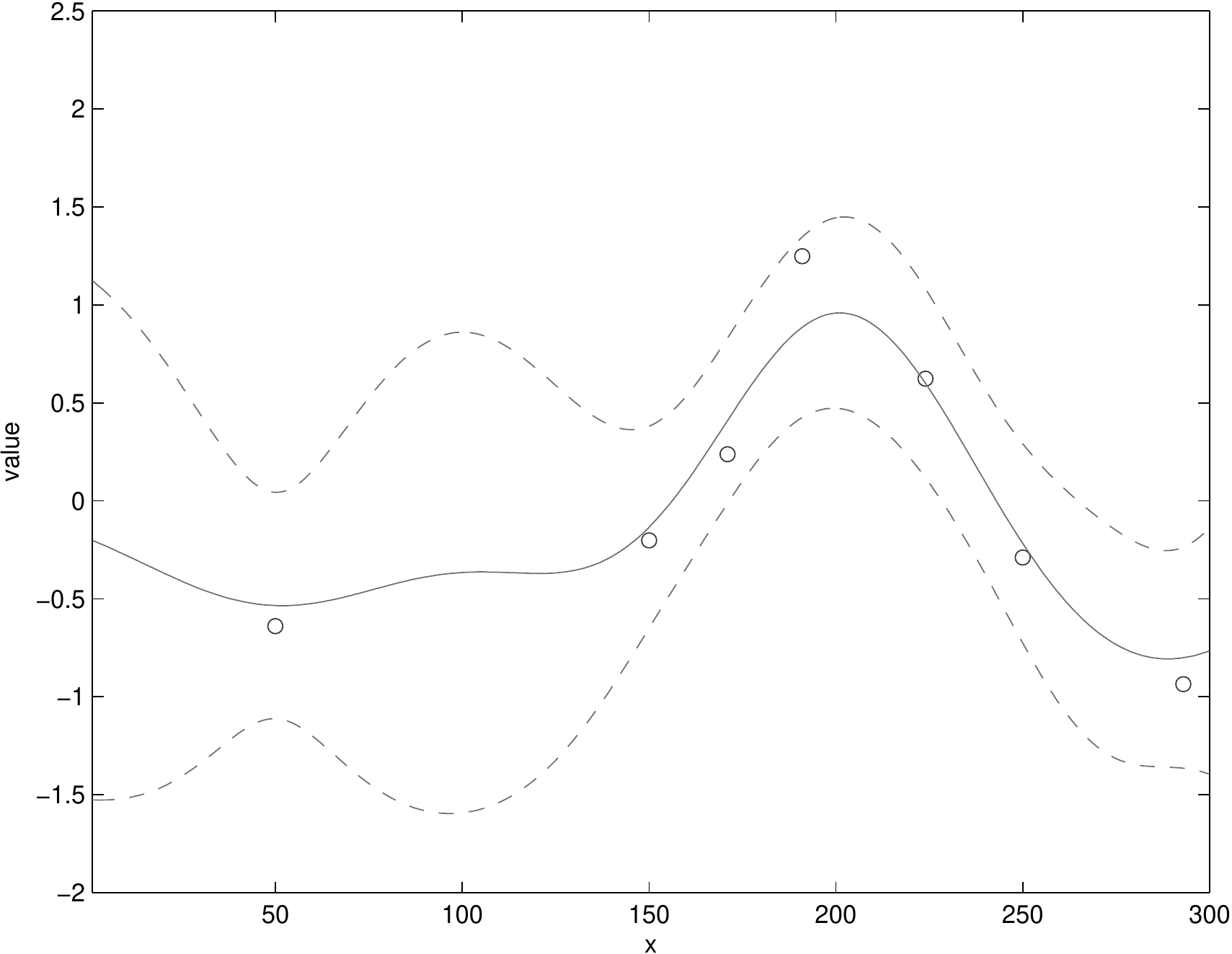}
\caption{Illustration of Gaussian process regression with noisy evaluations.  As in Figure~\ref{fig:GP-noise-free}, the circles show previously evaluated points, $(x_i,y_i)$, where $y_i$ is $f(x_i)$ perturbed by constant-variance independent noise.  The solid line shows the posterior mean, $\mu_n(x)$ as a function of $x$, which is an estimate of the underlying function $f$, and the dashed lines show a Bayesian credible interval for $f$, calculated as $\mu_n(x) \pm 1.96\sigma_n(x)$.
\label{fig:GP-noisy}}
    \end{center}
\end{figure}

Figure~\ref{fig:GP-noisy} shows an example of a posterior distribution calculated with Gaussian process regression with noisy observations.  Notice that the posterior mean no longer interpolates the observations, and the credible interval has a strictly positive width at points where we have measured.  Noise prevents us from observing function values exactly, and so we remain uncertain about the function value at points we have measured.

\optional{PF: need to make the axis legend bigger in Figure~\ref{fig:GP-noise-free} and \ref{fig:GP-noisy}}

\paragraph{Going beyond homoscedastic independent noise}

Constant variance is violated if the experimental noise differs across materials designs, which occurs most frequently when noise arises during the synthesis of the material itself, rather than during the evaluation of a material that has already been created.   Some work has been done to extend Gaussian process regression to flexibly model heteroscedastic noise (i.e., noise whose variance changes) \cite{GoWiBi98,KePlPfBu08,AnNeSt10,wang2014gaussian}.  The main idea in much of this work is to use a second Gaussian process to model the changing variance across the input domain.  Much of this work assumes that the noise is independent and Gaussian, though \cite{wang2014gaussian} considers non-Gaussian noise.

Independence is most typically violated, in the context of physical experiments, when the synthesis and evaluation of multiple materials designs is done together, and the variation in some shared component simultaneously influences these designs, e.g., through variation in the temperature while the designs are annealing together, or through variation in the quality of some constituent used in synthesis.  We are aware of relatively little work modeling dependent noise in the context of Gaussian process regression and Bayesian optimization, with one exception being \cite{FrazierXieChick2011}.

\subsection{Parameter Estimation}
\label{sec:parameter}
The mean and covariance functions contain several parameters.  For example, if we use the squared exponential kernel, a constant mean function, and observations have independent homoscedastic noise, then we must choose or estimate the parameters $\mu,\alpha,\beta_1,\ldots,\beta_d, \lambda$. These parameters are typically called {\it hyperparameters} because they are parameters of the prior distribution.  ($\lambda^2$ is actually a parameter of the likelihood function, but it is convenient to treat it together with the parameters of the prior.)  While one may simply choose these hyperparameters directly, based on intuition about the problem, a more common approach is to choose them adaptively, based on data.

To accomplish this, we write down an expression for the probability of the observed data $y_{1:n}$ in terms of the hyperparameters, marginalizing over the uncertainty on $f(x_{1:n})$.  Then, we optimize this expression over the hyperparameters to find settings that make the observed data as likely as possible.  This approach to setting hyperparameters is often called {\it empirical Bayes}, and it can be seen as an approximation to full Bayesian inference.

We detail this approach for the squared exponential kernel with a constant mean function.  Estimating for other kernels and mean functions is similar.
Using the probability distribution of $y_{1:n}$ from \eqref{eq:GP-prior-noise}, and neglecting constants, the natural logarithm of this probability, $\log p(y_{1:n} \mid x_{1:n})$ 
(called the ``log marginal likelihood''), can be calculated as
\begin{equation*}
\begin{split}
  -\frac{1}{2} (y_{1:n} - \mu)^T \left(\Sigma_0(x_{1:n}, x_{1:n}) + \lambda^2 I_n\right)^{-1} (y_{1:n} - \mu) - \frac{1}{2}\log |\Sigma_0(x_{1:n}, x_{1:n}) + \lambda^2 I_n|,
\end{split}
\end{equation*}
where $|\cdot|$ applied to a matrix indicates the determinant.

To find the hyperparameters that maximize this log marginal likelihood (the neglected constant does not affect the location of the maximizer), we will take partial derivatives with respect to each hyperparameter.  We will then use them to find maximizers of $\mu$ and $\sigma^2 := \alpha + \lambda^2$ analytically, and then use gradient-based optimization to maximize the other hyperparameters.

Taking a partial derivative with respect to $\mu$, setting it to zero, and solving for $\mu$, we get that the value of $\mu$ that maximizes the marginal likelihood is
\begin{equation*}
  \hat{\mu} = \frac{\sum_{i=1}^n \left((\Sigma_0(x_{1:n}, x_{1:n}) + \lambda^2 I_n)^{-1} y_{1:n}\right)_i}{\sum_{i,j=1}^n (\Sigma_0(x_{1:n}, x_{1:n}) + \lambda^2 I_n)^{-1}_{ij}}.
\end{equation*}
Define $R$ as the matrix with components
\begin{equation*}
  R_{ij} = \begin{dcases} 1 & i=j,\\ g \exp\left(-\sum_{i=1}^d \beta_i (x_i-x_j)^2\right) & i \neq j, \end{dcases}
\end{equation*}
where $g = \frac{\alpha}{\sigma^2}$. Then $\Sigma_0(x_{1:n}, x_{1:n}) + \lambda^2 I_n = \sigma^2 R$ and $\hat{\mu}$ can be written in terms of $R$ as $\hat{\mu} = \frac{\Sigma_{i=1}^n \left(R^{-1} y_{1:n}\right)_i}{\Sigma_{i,j=1}^n R^{-1}_{ij}}$.
The log marginal likelihood (still neglecting constants) becomes
\begin{equation*}
  \log p(y_{1:n} \mid x_{1:n}) \sim -\frac{1}{2}(y_{1:n}-\hat{\mu})^T(\sigma^2 R)^{-1}(y_{1:n}-\hat{\mu})-\frac{1}{2}\log|\sigma^2 R|.
\end{equation*}
Taking the partial derivative with respect to $\sigma^2$, and noting that $\hat{\mu}$ does not depend on $\sigma^2$, we solve for $\sigma^2$ and obtain
\begin{equation*}
  \widehat{\sigma^2} = \frac{1}{n}(y_{1:n}-\hat{\mu}) R^{-1} (y_{1:n} - \hat{\mu}).
\end{equation*}
Substituting this estimate, the log marginal likelihood becomes
\begin{equation}
  \log p(y_{1:n}\mid x_{1:n}) \sim -\log \left(\frac{1}{n} |R|^{\frac{1}{n}} (y_{1:n}-\hat{\mu})^TR^{-1}(y_{1:n}-\hat{\mu}) \right).
\label{eq:final_likelihood}
\end{equation}

The expression \eqref{eq:final_likelihood} cannot in general be optimized analytically.  Instead, one typically optimizes it numerically using a first- or second-order optimization algorithm, such as Newton's method or gradient descent, obtaining estimates for $\beta_1, \ldots, \beta_d$ and $g$.  These estimates are in turn substituted to provide an estimate of $R$, from which estimates $\hat{\mu}$ and $\widehat{\sigma^2}$ may be computed.  Finally, using $\widehat{\sigma^2}$ and the estimated value of $g$, we may estimate $\alpha$ and $\lambda$.

\subsection{Diagnostics}

\begin{figure}
\center
\includegraphics[scale=0.25]{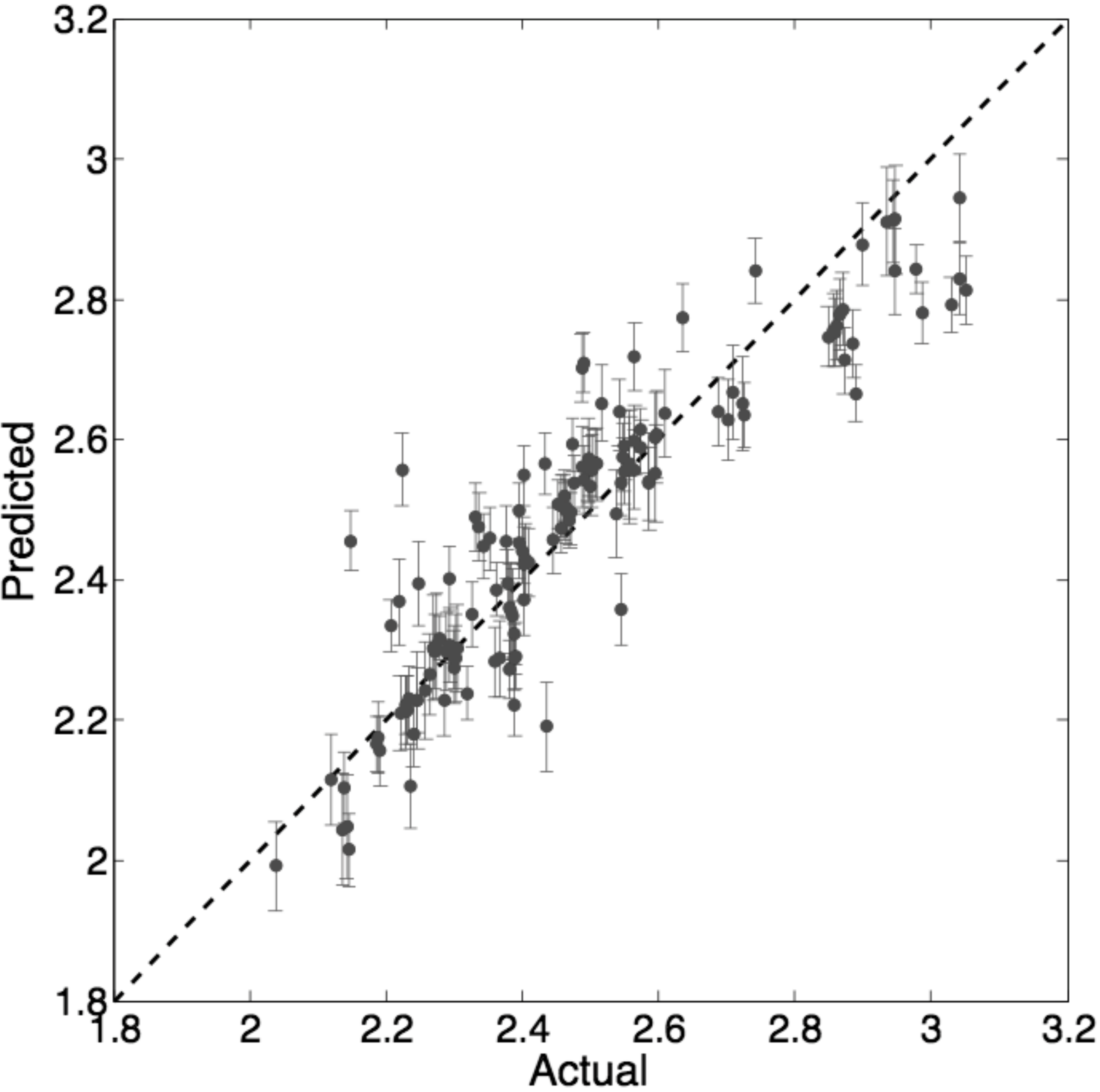}
\caption{Diagnostic plot for Gaussian process regression, created with leave-one-out cross validation.  For each point in our dataset, we hold that point $(x_i,y_i)$ out, train on the remaining points, calculate a 95\% credible interval for $y_i$, and plot this confidence interval as an error bar whose x-coordinate is the actual value $y_i$.  If Gaussian process regression is working well, 95\% of the error bars will intersect the diagonal line Predicted=Actual.
\label{fig:diagnostic}}
\end{figure}

When using Gaussian process regression, or any other machine learning technique, it is advisable to check the quality of the predictions, and to assess whether the assumptions made by the method are met.  One way to do this is illustrated by Figure~\ref{fig:diagnostic}, which comes from a simulation of blood flow near the heart, based on \cite{sankaran2010impact}, for which we get exact (not noisy) observations of $f(x)$

This plot is created with a technique called leave-one-out cross validation.  In this technique, we iterate through the datapoints $x_{1:n}$, $y_{1:n}$, and for each $i\in\{1,\ldots,n\}$, we train a Gaussian process regression model on all of the data {\it except} $x_i,y_i$, and then use it, together with $x_i$, to predict what the value $y_i$ should be.  We obtain from this a posterior mean (the prediction), call it $\mu_{-i}(x_i)$, and also a posterior standard deviation, call it $\sigma_{-i}(x_i)$. When calculating these estimates, it is best to separately re-estimate the hyperparameters each time, leaving out the data $(x_i,y_i)$.
We then calculate a 95\% credible interval $\mu_{-i}(x_i) \pm 2\sigma_{-i}(x_i)$, 
and create Figure~\ref{fig:diagnostic} by plotting ``Predicted'' vs. ``Actual'', where the 
``Actual'' coordinate (on the x-axis) is $y_i$, and the ``Predicted'' value (on the y-axis) is pictured as an error bar centered at $\mu_{-i}(x_i)$ with half-width $2\sigma_{-i}(x_i)$.

If the uncertainty estimates outputted by Gaussian process regression are behaving as anticipated, then 95\% of the credible intervals will intersect the diagonal line Predicted=Actual.  Moreover, if Gaussian process regression's predictive accuracy is high, then the credible intervals will be short, and their centers will be close to this same line Predicted=Actual.

This idea may be extended to noisy function evaluations, under the assumption of independent homoscedastic noise.  To handle the fact that the same point may be sampled multiple times, let $m(x)$ be the number of times that a point $x\in\{x_1,\ldots,x_n\}$ was sampled, and let $\ybar(x)$ be the average of the observed values at this point.  Moreover, by holding out all $m(x)$ samples of $x$ and training Gaussian process regression, we would obtain a normal posterior distribution on $f(x_i)$ that has mean $\mu_{-i}(x_i)$ and standard deviation $\sigma_{-i}(x_i)$.

Since $\ybar(x_i)$ is then the sum of $f(x_i)$ and some normally distributed noise with mean $0$ and variance $\lambda^2/m(x_i)$, the resulting distribution of $\ybar(x_i)$ is normal with mean $\mu_{-i}(x_i)$ and standard deviation $\sqrt{\sigma_{-i}^2(x_i) + \lambda^2/m(x_i)}$.

From this, a 95\% credible interval for $\ybar(x_i)$ is then $\mu_{-i}(x_i) \pm 2 \sqrt{\sigma_{-i}^2(x_i) + \lambda^2/m(x_i)}$.  We would plot Predicted vs. Observed by putting this credible interval along the y-axis at x-coordinate $\ybar(x_i)$.  If Gaussian process regression is working well, then 95\% of these credible intervals will intersect the line Predicted=Observed.

For Gaussian process regression to best support Bayesian optimization, it is typically most important to have good uncertainty estimates, and relatively less important to have high predictive accuracy.  This is because Bayesian optimization uses Gaussian process regression as a guide for deciding where to sample, and so if Gaussian process regression reports that there is a great deal of uncertainty at a particular location and thus low predictive accuracy, Bayesian optimization can choose to sample at this location to improve accuracy.  Thus, Bayesian optimization has a recourse for dealing with low predictive accuracy, as long as the uncertainty is accurately reported.
In contrast, if Gaussian process regression estimates poor performance at a location that actually has near-optimal performance, and also provides an inappropriately low error estimate, then Bayesian optimization may not sample there within a reasonable timeframe, and thus may never correct the error.

If either the uncertainty is incorrectly estimated, or the predictive accuracy is unsatisfactorily low, then the most common ``fixes'' employed are to adopt a different covariance kernel, or to transform the objective function $f$.  If the objective function is known to be non-negative, then the transformations $\log(f)$ and $\sqrt{f}$ are convenient for optimization because they are both strictly increasing, and so do not change the set of maximizers (or minimizers).  If $f$ is not non-negative, but is bounded below by some other known quantity $a$, then one may first shift $f$ upward by $a$.

\subsection{Predicting at more than one point}
Below, to support the development of the knowledge-gradient method in Sections~\ref{sec:KG} and~\ref{sec:KGcalculation}, it will be useful to predict the value of $f$ at multiple points, $x^*_1,\ldots,x^*_k$, with noise.
To do so, we could certainly apply \eqref{eq:GP-noisy-posterior-mean} and \eqref{eq:GP-noisy-posterior-covariance} separately for each $x^*_1,\ldots,x^*_k$, and this would provide us with both an estimate (the posterior mean) and a measure of the size of the error in this estimate (the posterior variance) associated with each $f(x^*_i)$.  It would not, however, quantify the relationship between the errors at several different locations.  For this, we must perform the estimation jointly.

As we did in Section~\ref{sec:GP-noise}, we begin with our prior on $[y_{1:n},f(\xpred_{1:k})]$, which is,
\begin{equation*}
\begin{bmatrix}
y_{1:n}\\ f(\xpred_{1:k})
\end{bmatrix}
\sim
\Normal\left(
\begin{bmatrix}
\mu_0(x_{1:n})\\ \mu_0(\xpred_{1:k})
\end{bmatrix},
\begin{bmatrix}
\Sigma_0(x_{1:n},x_{1:n})+\lambda^2 I_n & \Sigma_0(x_{1:n},\xpred_{1:k}) \\
\Sigma_0(\xpred_{1:k},x_{1:n})                & \Sigma_0(\xpred_{1:k},\xpred_{1:k})
\end{bmatrix}
\right),
\end{equation*}

We then use Proposition~\ref{prop:multivariate-normal-conditioning} to compute the posterior on $f(\xpred_{1:k})$ given $f(x_{1:n})$, which is multivariate normal with mean vector $\mu_n(\xpred_{1:k})$ and covariance matrix $\Sigma_n(\xpred_{1:k},\xpred_{1:k})$ given by,
\begin{align}
\mu_n(\xpred_{1:k})&=\mu_0(\xpred_{1:k})+\Sigma_0(\xpred_{1:k},x_{1:n})\left[\Sigma_0(x_{1:n},x_{1:n})+\lambda^2 I_n\right]^{-1}(y_{1:n}-\mu_0(x_{1:n})),
\label{eq:GP-noisy-posterior-mean-multiple} \\
\Sigma_n(\xpred_{1:k},\xpred_{1:k})&=\Sigma_0(\xpred_{1:k},\xpred_{1:k})-\Sigma_0(\xpred_{1:k},x_{1:n})\left[\Sigma_0(x_{1:n},x_{1:n})+\lambda^2 I_n\right]^{-1}\Sigma_0(x_{1:n},\xpred_{1:k}).
\label{eq:GP-noisy-posterior-covariance-multiple}
\end{align}

We see that setting $k=1$ provides the expressions \eqref{eq:GP-noisy-posterior-mean} and \eqref{eq:GP-noisy-posterior-covariance} from Section~\ref{sec:GP-noise}.

\subsection{Avoiding matrix inversion}
The expressions \eqref{eq:GP-noisy-posterior-mean} and \eqref{eq:GP-noisy-posterior-covariance} for the posterior mean and variance in the noisy case, and also \eqref{eq:GP-intuition-mean} and \eqref{eq:GP-intuition-variance} in the noise-free case,
include a matrix inversion term.  Calculating this matrix inversion is slow and can be hard to accomplish accurately in practice, due to the finite precision of floating point implementations. Accuracy is especially an issue when $\Sigma$ has terms that are close to $0$, which arises when data points are close together.

In practice, rather than calculating a matrix inverse directly, it 
is typically faster and more accurate to 
use a mathematically equivalent algorithm, which performs a Cholesky decomposition and then solves a linear system.  This algorithm is described below, and is adapted from Algorithm~2.1 in Section~2.3 of \cite{RaWi06}.
This algorithm also computes the log marginal likelihood required for estimating hyperparameters in Section~\ref{sec:parameter}.
\begin{algorithm}
\caption{Implementation using Cholesky decomposition}\label{algo:chol}
\begin{algorithmic}[1]
  \REQUIRE $x_{1:n}$ (inputs), $y_{1:n}$ (responses), $\Sigma_0(x, x')$ (covariance function), $\lambda^2$ (variance of noise), $\xpred$ (test input).
  \STATE $L = \text{Cholesky}\left(\Sigma_0(x_{1:n}, x_{1:n}) + \lambda^2 I_n\right)$
  \STATE $\delta = L^T \backslash \left(L \backslash \left( y_{1:n} - \mu_0(x_{1:n})\right)\right)$
  \STATE $\mu_n(\xpred) = \mu_0(\xpred) + \Sigma_0(\xpred, x_{1:n}) \delta$
  \STATE $v = L \backslash \Sigma_0(x_{1:n}, \xpred)$
  \STATE $\sigma_n^2(\xpred) = \Sigma_0(\xpred, \xpred) - v^T v$
  \STATE $\log p(y_{1:n} \mid x_{1:n}) = -\frac{1}{2} \left( y_{1:n} - \mu_0(x_{1:n}) \right)^T \alpha - \Sigma_i \log L_{ii} - \frac{n}{2}\log 2\pi$
  \RETURN $\mu_n(\xpred)$ (mean), $\sigma_n^2(\xpred)$ (variance), $\log p(y_{1:n} \mid x_{1:n})$ (log  marginal likelihood).
\end{algorithmic}
\end{algorithm}

\section{Choosing where to sample}
\label{sec:recommendation}
Being able to infer the value of the objective function $f(x)$ at unevaluated points based on past data $x_{1:n}$,$y_{1:n}$ is only one part of finding good designs.  The other part is using this ability to make good decisions about where to direct future sampling.

Bayesian optimization methods addresses this by using a measure of the value of the information that would be gained by sampling at a point.  Bayesian optimization methods then choose the point to sample next for which this value is largest.  A number of different ways of measuring the value of information have been proposed.  Here, we describe two in detail, expected improvement \cite{Mo89,JoScWe98}, and the knowledge gradient \cite{FrazierPowellDayanik2009,ScottFrazierPowell2011}, and then survey a broader collection of design criteria.

\subsection{Expected Improvement}
\label{sec:EI}
Expected improvement, as it was first proposed, considered only the case where measurements are free from noise.
In this setting, suppose we have taken $n$ measurements at locations $x_{1:n}$ and observed $y_{1:n}$.  Then
\begin{equation*}
f^*_n = \max_{i=1,\ldots,n} f(x_i)
\end{equation*}
is the best value observed so far.  Suppose we are considering evaluating $f$ at a new point $x$.
After this evaluation, the best value observed will be 
\begin{equation*}
f^*_{n+1} = \max( f(x), f^*_n),
\end{equation*}
and the difference between these values, which is the {\it improvement} due to sampling, is
\begin{equation*}
f^*_{n+1} - f^*_n = \max(f(x) - f^*_n, 0) = (f(x)-f^*_n)^+,
\end{equation*}
where $a^+ = \max(a,0)$ indicates the positive part function.

Ideally, we would choose $x$ to make this improvement as large as possible.  Before actually evaluating $f(x)$, however, we do not know what this improvement will be, so we cannot implement this strategy.  However, we do have a probability distribution on $f(x)$, from Gaussian process regression.  The {\it expected improvement}, indicated $\EI(x)$, is obtained by taking the expectation of this improvement with respect to the posterior distribution on $f(x)$ given $x_{1:n},y_{1:n}$.
\begin{equation}
\EI(x) = E_n[ (f(x)-f^*_n)^+], \label{eq:EI}
\end{equation}
where $E_n[\ \cdot\ ] = E[\ \cdot\ | x_{1:n}, y_{1:n}]$ indicates the expectation with respect to the posterior distribution.

\begin{figure}
\center
\includegraphics[scale=0.6]{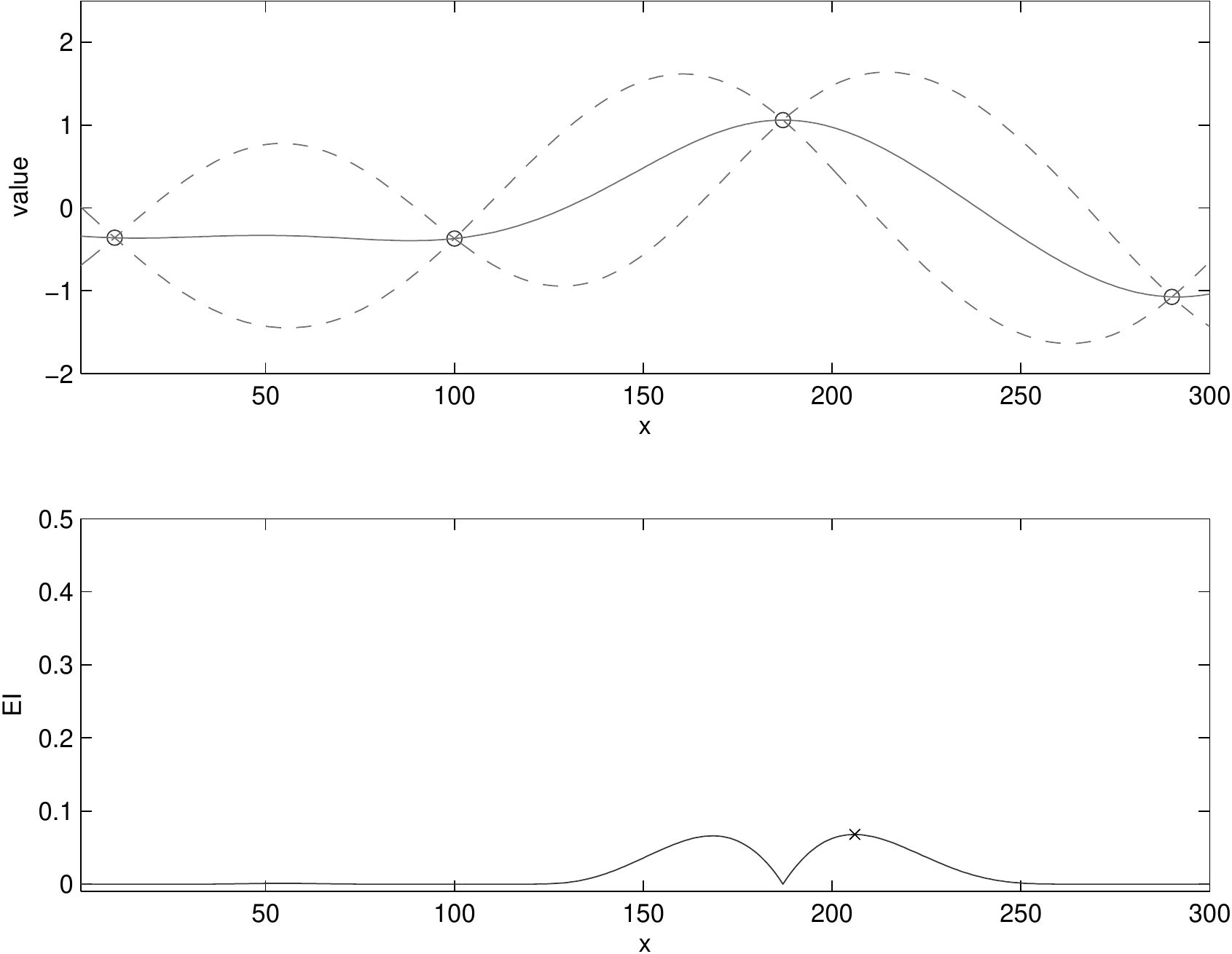}
\caption{Upper panel shows the posterior distribution in a problem with no noise and a one-dimensional input space, where the circles are previously measured points, the solid line is the posterior mean $\mu_n(x)$, and the dashed lines are at $\mu_n(x) \pm 2\sigma_n(x)$.
Lower panel shows the expected improvement $\EI(x)$ computed from this posterior distribution.  An ``x'' is marked at the point with the largest expected improvement, which is where we would evaluate next.
\label{fig:EI}}
\end{figure}

The expectation in \eqref{eq:EI} can be computed more explicitly, in terms of the normal cumulative distribution function (cdf) $\Phi(\cdot)$, and the normal probability density function (pdf) $\varphi(\cdot)$.  
Recalling from Section~\ref{sec:inference} that $f(x) \sim \Normal(\mu_n(x),\sigma^2_n(x))$, where $\mu_n(x)$ and $\sigma^2_n(x)$ are given by \eqref{eq:GP-noise-free-posterior-mean} and \eqref{eq:GP-noise-free-posterior-covariance}, and integrating with respect to the normal distribution (a derivation may be found in Section~\ref{sec:derivations}), we obtain,
\begin{equation}
\label{eq:EI_analytic}
\EI(x) = 
(\mu_n(x) - f^*_n)\Phi\left(\frac{\mu_n(x) - f^*_n}{\sigma_n(x)}\right) +
\sigma_n(x) \varphi\left(\frac{\mu_n(x) - f^*_n}{\sigma_n(x)}\right).
\end{equation}
Figure~\ref{fig:EI} plots this expected improvement for a problem with a one-dimensional input space.
We can see from this plot that the expected improvement is largest at locations where both the posterior mean $\mu_n(x)$ is large, and also the posterior standard deviation $\sigma^n_x$ is large.  This is reasonable because those points that are most likely to provide large gains are those points that have a high predicted value, but that also have significant uncertainty.  Indeed, at points where we have already observed, and thus have no uncertainty, the expected improvement is $0$.  This is consistent with the idea that, in a problem without noise, there is no value to repeating an evaluation that has already been performed.

This idea of favoring points that, on the one hand, have a large predicted value, but, on the other hand, have a significant amount of uncertainty, is called the {\it exploration vs. exploitation tradeoff}, and appears in areas beyond Bayesian optimization, especially in reinforcement learning \cite{Ka93,SuBa98} and multi-armed bandit problems \cite{Gittins2011,MaTe07}.  In these problems, we are taking actions repeatedly over time whose payoffs are uncertain, and wish to simultaneously get good immediate rewards, while learning the reward distributions for all actions to allow us to get better rewards in the future.
We emphasize, however, that the correct balance between exploration and exploitation is different in Bayesian optimization as compared with multi-armed bandits, and should more favor exploration: in optimization, the advantage of measuring where the predicted value is high is that these areas tend to give more useful information about where the optimum lies; in contrast, in problems where we must ``learn while doing'' like multi-armed bandits, evaluating an action with high predicted reward is good primarily because it tends to give a high immediate reward.

\begin{figure}
\center
\includegraphics[scale=0.35]{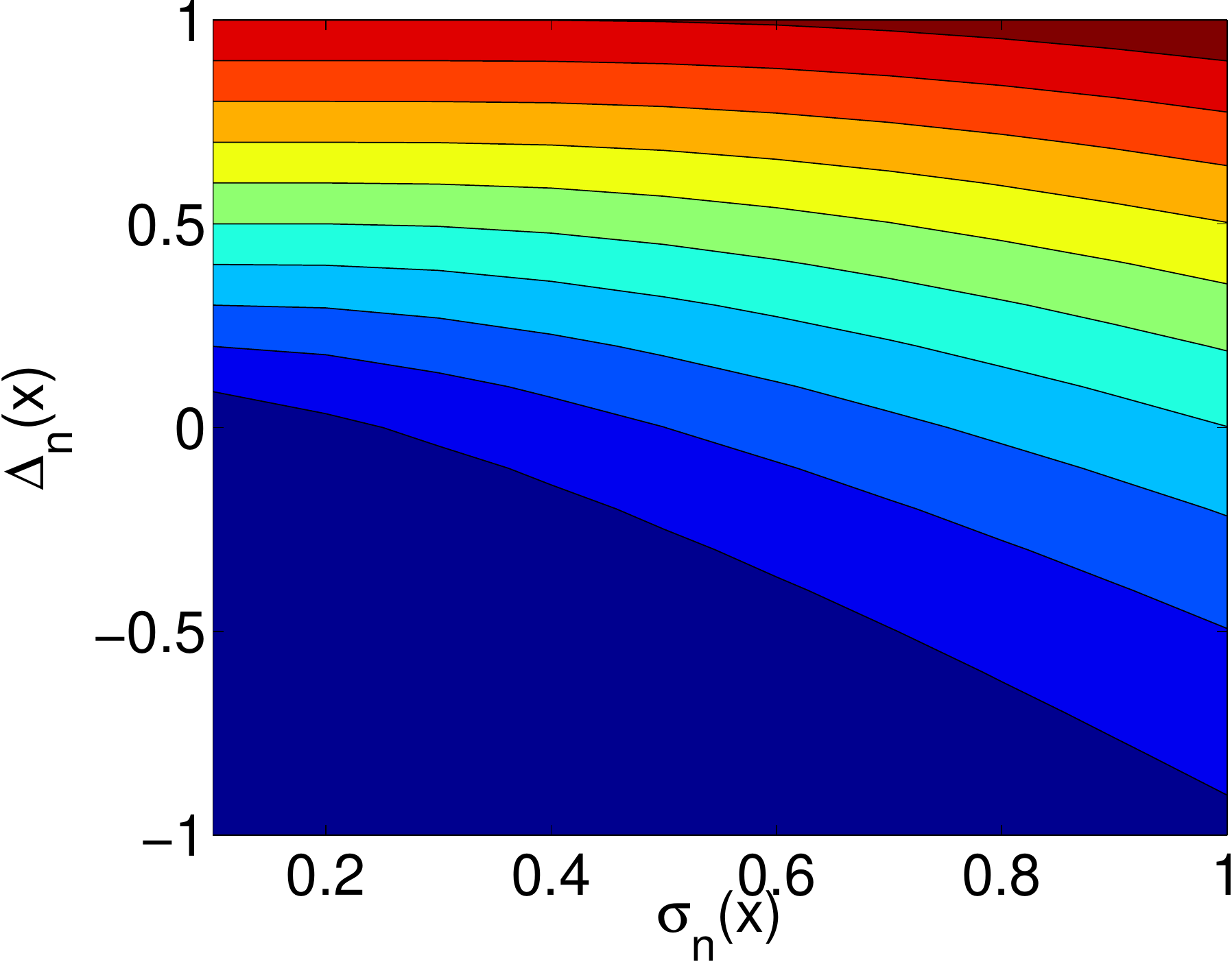}
\caption{Contour plot of the expected improvement, as a function of the difference in means $\Delta_n(x) := \mu_n(x) - f^*_n$ and the posterior standard deviation $\sigma_n(x)$.
The expected improvement is larger when the difference in means is larger, and when the standard deviation is larger.
\label{fig:EIcontour}}
\end{figure}

We can also see the exploration vs. exploitation tradeoff implicit in the expected improvement function in the contour plot, Figure~\ref{fig:EIcontour}.  This plot shows the contours of $\EI(x)$ as a function of the posterior mean, expressed as a difference from the previous best, $\Delta_n(x) := \mu_n(x) - f^*_n$, and the posterior standard deviation $\sigma_n(x)$.

Given the expression~\eqref{eq:EI_analytic}, Bayesian optimization algorithms based on expected improvement, such as the Efficient Global Optimization (EGO) algorithm proposed by \cite{JoScWe98}, and the earlier algorithms of Mockus (see, e.g., the monograph \cite{Mo89}), then recommend sampling at the point with the largest expected improvement.  That is,
\begin{equation}
\label{eq:EI_optimization}
x_{n+1} \in \argmax_x \EI(x).
\end{equation}

Finding the point with largest expected improvement is itself a global optimization problem, like the original problem that we wished to solve \eqref{eq:obj}.  Unlike \eqref{eq:obj}, however, $\EI(x)$ can be computed quickly, and its first and second derivatives can also be computed quickly.  Thus, we can expect to be able to solve \eqref{eq:obj} relatively well using an off-the-shelf optimization method for continuous global optimization.  A common approach is to use a local solver for continuous optimization, such as gradient ascent, in a multistart framework, where we start the local solver from many starting points chosen at random, and then select the best local solution discovered.  In Section~\ref{sec:software} we describe several codes that implement expected improvement methods, and each makes its own choice about how to solve \eqref{eq:EI_optimization}.

The algorithm given by \eqref{eq:EI_optimization} is {\it optimal} under three assumptions:
(1) that we will take only a single sample;
(2) there is no noise in our samples; and
(3) that the $x$ we will report as our final solution (i.e., the one that we will {\it implement}) must be among those previously sampled.

In practice, assumption (1) is violated, as Bayesian optimization methods like \eqref{eq:EI_optimization} are applied iteratively, and is made simply because it simplifies the analysis.  Being able to handle violations of assumption (1) in a more principled way is of great interest to researchers working on Bayesian optimization methodology, and some partial progress in that direction is discussed in Section~\ref{sec:other-methods}.
Assumption (2) is also often violated in a broad class of applications, especially those involving physical experiments or stochastic simulations.  In the next section, we present an algorithm, the knowledge-gradient algorithm \cite{FrazierPowellDayanik2009,ScottFrazierPowell2011}, that relaxes this assumption (2), and also allows relaxing assumption (3) if this is desired.

\subsection{Knowledge Gradient}
\label{sec:KG}
When we have noise in our samples, the derivation of expected improvement meets with difficulty.
In particular, if we have noise, then $f^*_n = \max_{i=1,\ldots,n} f(x_i)$ 
is not precisely known, preventing us from using the expression \eqref{eq:EI_analytic}.

One may simply take a quantity like $\max_{i=1,\ldots,n} y_i$
that is similar in spirit to 
$f^*_n = \max_{i=1,\ldots,n} f(x_i)$, and replace $f^*_n$ in \eqref{eq:EI_analytic} with this quantity, but the resulting algorithm is no longer justified by an optimality analysis.  Indeed, for problems with a great deal of noise, 
$\max_{i=1,\ldots,n} y_i$ tends to be significantly larger than the true underlying value of the best point previously sampled, and so the resulting algorithm may be led to make a poor tradeoff between exploration and exploitation, and exhibit poor performance in such situations.

Instead, the knowledge-gradient algorithm \cite{FrazierPowellDayanik2009,ScottFrazierPowell2011} takes a more principled approach, and starts where the derivation of expected improvement began, but fully accounts for the introduction of noise (assumption 2 in section~\ref{sec:EI}), and the possibility that we wish to search over a class of solutions broader than just those that have been previously evaluated when recommending the final solution (assumption 3 in section~\ref{sec:EI}).

We first introduce a set $A_n$, which is the set of points from which we would choose final solution, if we were asked to recommend a final solution at time $n$, based on $x_{1:n}$, $y_{1:n}$.
For tractability, we suppose $A_n$ is finite.
For example, if $A$ is finite, as it often is in discrete optimization via simulation problems, we could take $A_n=A$, allowing the whole space of feasible solutions.  This choice was considered in \cite{FrazierPowellDayanik2009}.  Alternatively, one could take $A_n=\{x_1,\ldots,x_n\}$, stating that one is willing to consider only those points that have been previously evaluated.  This choice is consistent with the expected improvement algorithm.  Indeed, we will see that when one makes this choice, and measurements are free from noise, then the knowledge-gradient algorithm is identical to the expected improvement algorithm.  Thus, the knowledge-gradient algorithm generalizes the expected improvement algorithm.


If we were to stop sampling at time $n$, then the expected value of a point $x\in A_n$ based on the information available would be $E_n[f(x)] = \mu_n(x)$.  In the special case when evaluations are free from noise, this is equal to $f(x)$, but when there is noise, these two quantities may differ.  If we needed to report a final solution, we would then choose the point in $A_n$ for which this quantity is the largest, i.e., we would choose $\argmax_{x\in A_n} \mu_n(x)$.  Moreover, the expected value of this solution would be
\begin{equation*}
\mu^*_n = \max_{x\in A_n} \mu_n(x).
\end{equation*}
If evaluations are free from noise and $A_n = \{x_{1:n}\}$, then $\mu^*_n$ is equal to $f^*_n$, but in general these quantities may differ.

If we take one additional sample, then the expected value of the solution we would report based on this additional information is
\begin{equation*}
\mu^*_{n+1} = \max_{x\in A_{n+1}} \mu_{n+1}(x),
\end{equation*}
where as before, $A_{n+1}$ is some finite set of points we would be willing to consider when choosing a final solution.
Observe in this expression that $\mu_{n+1}(x)$ is not necessarily the same as $\mu_{n}(x)$, even for points $x \in \{x_{1:n}\}$ that we had previously evaluated, but that $\mu_{n+1}(x)$ can be computed from the history of observations $x_{1:n+1}$, $y_{1:n+1}$.

The improvement in our expected solution value is then the difference between these two quantities, $\mu^*_{n+1} - \mu^*_n$.  This improvement is random at time $n$, even fixing $x_{n+1}$, through its dependence on $y_{n+1}$, but we can take its expectation.  The resulting quantity is called the {\it knowledge-gradient (KG) factor}, and is written,
\begin{equation}
\KG_n(x) = E_n\left[ \mu^*_{n+1} - \mu^*_n \mid x_{n+1}=x \right].
\label{eq:KG}
\end{equation}

Calculating this expectation is more involved than calculating the expected improvement, but nevertheless can also be done analytically in terms of the normal pdf and normal cdf.  This is described in more detail in Section~\ref{sec:KGcalculation}.

The knowledge-gradient algorithm is then the one that chooses the point to sample next that maximizes the KG factor,
\begin{equation*}
\argmax_x \KG_n(x).
\end{equation*}

\begin{figure}
\center
\includegraphics[scale=0.6]{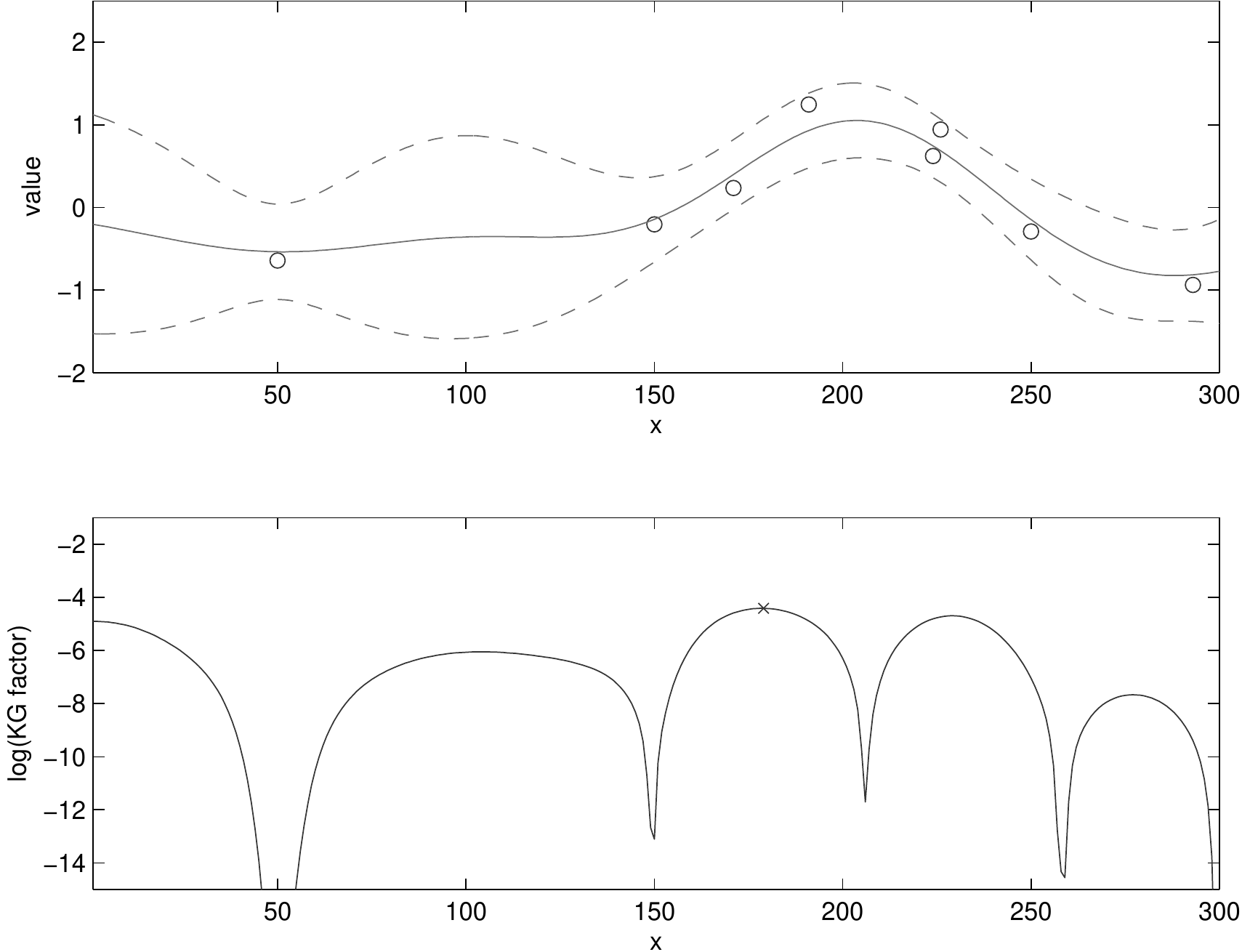}
\caption{Upper panel shows the posterior distribution in a problem with independent normal homoscedastic noise and a one-dimensional input space, where the circles are previously measured points, the solid line is the posterior mean $\mu_n(x)$, and the dashed lines are at $\mu_n(x) \pm 2\sigma_n(x)$.
Lower panel shows the natural logarithm of the knowledge-gradient factor $\KG(x)$ computed from this posterior distribution, where $A_n=A_{n+1}$ are the discrete grid $\{1,\ldots,300\}$.  An ``x'' is marked at the point with the largest KG factor, which is where the KG algorithm would evaluate next.
\label{fig:KG}}
\end{figure}

The KG factor for a one-dimensional optimization problem with noise is pictured in Figure~\ref{fig:KG}.
We see a similar tradeoff between exploration and exploitation, where the KG factor favors measuring points with a large $\mu_n(x)$ and a large $\sigma_n(x)$.  We also see local minima of the KG factor at points where we previously evaluated, just as with the expected improvement, but because there is noise in our samples, the value at these points is not $0$ --- indeed, when there is noise, it may be useful to sample repeatedly at a point.

\paragraph{Choice of $A_n$ and $A_{n+1}$}

Recall that the KG factor depends on the choice of the sets $A_n$ and $A_{n+1}$, through the dependence of $\mu^*_n$ and $\mu^*_{n+1}$ on these sets. Typically, if we choose these sets to contain more elements, then we allow $\mu^*_n$ and $\mu^*_{n+1}$ to range over a larger portion of the space, and we allow the KG factor calculation to more accurately approximate the value that would result if we allowed ourself to implement the best option.  However, as we increase the size of these sets, computing the KG factor is slower, making implementation of the KG method more computationally intensive.

For applications with a finite $A$, \cite{FrazierPowellDayanik2009} proposed setting $A_{n+1}=A_n=A$, which was seen to require fewer function evaluations to find points with large $f$,  in comparison with expected improvement on noise-free problems, and in comparison with another Bayesian optimization method, sequential kriging optimization (SKO) \cite{HuAlNoZe06} on noisy problems.  However, the computation and memory required grows rapidly with the size of $A$, and is typically not feasible when $A$ contains more than 10,000 points.

For large-scale applications, \cite{ScottFrazierPowell2011} proposed setting $A_{n+1} = A_n = \{x_{1:n+1}\}$ in \eqref{eq:KG}, and called the resulting quantity the {\it approximate knowledge gradient (AKG)}, observing that this choice maintained computational tractability as $A$ grows, but also offers good performance. This algorithm is implemented in the DiceKriging package \cite{roustant2012dicekriging}.

Finally, in noise-free problems (but not in problems with noise), setting $A_{n+1} = \{x_{1:n+1}\}$ and $A_n = \{x_{1:n}\}$ recovers expected improvement.

\subsection{Going beyond one-step analyses, and other methods}
\label{sec:other-methods}

Both expected improvement and the knowledge-gradient method are designed to be optimal, in the special case where we will take just one more function evaluation and then choose a final solution.  They are not, however, known to be optimal for the more general case in which we will take multiple measurements, which is the way they are used in practice.  

The optimal algorithm for this more general setting is understood to be the solution to a partially observable Markov decision process, but actually computing the optimal solution using this understanding is intractable using current methods \cite{Frazier2011a}.  Some work has been done toward the goal of developing such an optimal algorithm \cite{GinsbourgerRiche2010}, but computing the optimal algorithm remains out of reach.  Optimal strategies have been computed for other closely related problems in optimization of expensive noisy functions, 
including stochastic root-finding \cite{WaeberFrazierHenderson2013}, multiple comparisons with a standard \cite{XieFrazier2013a}, and small instances of discrete noisy optimization with normally distributed noise (also called ``ranking and selection'') \cite{Frazier2012}.

Expected improvement and the knowledge gradient are both special cases of the more general concept of value of information, or expected value of sample information (EVSI) \cite{Ho66}, as they calculate the expected reward of a final implementation decision as a function of the posterior distribution resulting from some information, subtract from this the expected reward that would result from not having the information, and then take the expectation of this difference with respect to the information itself.

Many other Bayesian optimization methods have been proposed.  A few of these methods optimize the value of information, but are calculated using different assumptions than those used to derive expected improvement or value of information.
A larger number of these methods optimize quantities that do not correspond to a value of information, but are derived using analyses that are similar in spirit.
These include methods that optimize the probability of improvement \cite{Ku64,St88,Pe91},
the entropy of the posterior distribution on the location of the maximum \cite{ViVaWa08},
and other composite measures involving the mean and the standard deviation of the posterior \cite{HuAlNoZe06}.

Other Bayesian optimization methods are designed for problem settings that do not match the assumptions made in this tutorial.  
These include \cite{Ba09,FrazierKazachkov2011,Kn06}, which consider multiple objectives;
\cite{GiLeCa08,GiLeCa10,ClarkWang,snoek2012practical}, which consider multiple simultaneous function evaluations;
\cite{FrazierPowellSimao2009,HuAlNoMi06,Forrester2007}, which consider objective functions that can be evaluated with multiple fidelities and costs;
\cite{BeGiLiPiVa10}, which considers Bernoulli outcomes, rather than normally distributed ones;
\cite{gardner2014bayesian}, which considers expensive-to-evaluate inequality constraints;
and \cite{FrazierNegoescuPowell2011}, which considers optimization over the space of small molecules.

\section{Software}
\label{sec:software}
There are a number of excellent software packages, both freely available and commercial, that implement the methods described in this chapter, and other similar methods.

\begin{itemize}
\item Metrics Optimization Engine (MOE), an open-source code in C++ and Python, developed by the authors and engineers at Yelp. \url{http://yelp.github.io/MOE/},
\item Spearmint, an open-source code in Python, implementing algorithms described in \cite{snoek2012practical}.
\url{https://github.com/JasperSnoek/spearmint}
\item DiceKriging and DiceOptim, an open-source R package that implements expected improvement, the approximate knowledge-gradient method, and a variety of algorithms for parallel evaluations. An overview is provided in \cite{roustant2012dicekriging}.\\
\url{http://cran.r-project.org/web/packages/DiceOptim/index.html},
\item TOMLAB, a commercial package for MATLAB. \url{http://tomopt.com/tomlab/}
\item matlabKG, an open-source research code that implements the discrete knowledge-gradient method for small-scale problems.\\
\url{http://people.orie.cornell.edu/pfrazier/src.html}
\end{itemize}

A list of software packages focused on Gaussian process regression (but not Bayesian optimization) may be found at \url{http://www.gaussianprocess.org/}.

\section{Conclusion}
\label{sec:conclusion}
We have presented Bayesian optimization, including Gaussian process regression, the expected improvement method, and the knowledge-gradient method.
In making this presentation, we wish to emphasize that this approach to materials design acknowledges the inherent uncertainty in statistical prediction and seeks to guide experimentation in a way that is robust to this uncertainty.  It is inherently iterative, and does not seek to circumvent the fundamental trial-and-error process.

This is in contrast with another approach to informatics in materials design, which holds the hope that predictive methods can short-circuit the iterative loop entirely.  In this alternative view of the world, one hopes to create extremely accurate prediction techniques, either through physically-motivated {\it ab initio} calculations, or using data-driven machine learning approaches, that are so accurate that one can rely on the predictions alone rather than on physical experiments.  If this can be achieved, then we can search over materials designs {\it in silico}, find those designs that are predicted to perform best, and test those designs alone in physical experiments.

For this approach to be successful, one must have extremely accurate predictions, which limits its applicability to settings where this is possible.  We argue that, in contrast, predictive techniques can be extremely powerful even if they are not perfectly accurate, as long as they are used in a way that acknowledges inaccuracy, builds in robustness, and reduces this inaccuracy through an iterative dialog with physical reality mediated by physical experiments.  Moreover, we argue that mathematical techniques like Bayesian optimization, Bayesian experimental design, and optimal learning provide us the mathematical framework for accomplishing this goal in a principled manner, and for using our power to predict as effectively as possible.

\section{Acknowledgements}
Peter I. Frazier was supported by AFOSR FA9550-12-1-0200, AFOSR FA9550-15-1-0038, NSF CAREER CMMI-1254298, NSF IIS-1247696, and the ACSF's AVF. Jialei Wang was supported by AFOSR FA9550-12-1-0200.

\section{Derivations and Proofs}\label{sec:derivations}
This section contains derivations and proofs of equations and theoretical results found in the main text.

\subsection{Proof of Proposition~\ref{prop:multivariate-normal-conditioning}}
\begin{proof}
Using Bayes' rule, the conditional probability density of $\theta_{[2]}$ at a point $u_{[2]}$ given that $\theta_{[1]} = u_{[1]}$ is
\begin{align}
  p(\theta_{[2]} = u_{[2]} &\mid \theta_{[1]} = u_{[1]}) 
  = \frac{p(\theta_{[1]} = u_{[1]}, \theta_{[2]}=u_{[2]})}{p(\theta_{[1]} = u_{[1]})}
  \propto p(\theta_{[1]} = u_{[1]}, \theta_{[2]}=u_{[2]}) \notag \\
  &\propto \exp \left( -\frac{1}{2} \begin{bmatrix} u_{[1]} - \mu_{[1]} \\ u_{[2]} - \mu_{[2]}\end{bmatrix}^T \begin{bmatrix}\Sigma_{[1,1]} & \Sigma_{[1,2]} \\ \Sigma_{[2,1]} & \Sigma_{[2,2]}\end{bmatrix}^{-1} \begin{bmatrix} u_{[1]} - \mu_{[1]} \\ u_{[2]} - \mu_{[2]}\end{bmatrix} \right).
  \label{eq:conditional_prob}
\end{align}
To deal with the inverse matrix in this expression, we use the following identity for inverting a block matrix: the inverse of the block matrix $\begin{bmatrix} A & B \\ C & D\end{bmatrix}$, where both $A$ and $D$ are invertible square matrices, is
\begin{equation}
  \begin{bmatrix} A & B \\ C & D\end{bmatrix}^{-1} = \begin{bmatrix} (A - BD^{-1}C)^{-1} & -(A-BD^{-1}C)^{-1}BD^{-1} \\ -(D-CA^{-1}B)^{-1}CA^{-1} & (D-CA^{-1}B)^{-1}\end{bmatrix}.
  \label{eq:block_matrix_inverse}
\end{equation}
Applying \eqref{eq:block_matrix_inverse} to \eqref{eq:conditional_prob}, and using a bit of algebraic manipulation to get rid of constants, we have
\begin{equation}
  p(\theta_{[2]}=u_{[2]} \mid \theta_{[1]} = u_{[1]}) \propto \exp \left( -\frac{1}{2}(u_{[2]} - \mu^{\text{new}})^T (\Sigma^{\text{new}})^{-1} (u_{[2]} - \mu^{\text{new}}) \right),
  \label{eq:conditional_prob_final}
\end{equation}
where $\mu^{\text{new}} = \mu_{[2]} - \Sigma_{[2,1]}\Sigma_{[1,1]}^{-1}(u_{[1]} - \mu_{[1]})$ and $\Sigma^{\text{new}} = \Sigma_{[2,2]}-\Sigma_{[2,1]}\Sigma_{[1,1]}^{-1}\Sigma_{[1,2]}$. 

We see that \eqref{eq:conditional_prob_final} is simply the unnormalized probability density function of a normal distribution.  Thus the conditional distribution of $\theta_{[2]}$ given $\theta_{[1]} = u_{[1]}$ is multivariate normal, with mean $\mu^{\text{new}}$ and covariance matrix $\Sigma^{\text{new}}$.
\end{proof}

\subsection{Derivation of Equation~\eqref{eq:EI_analytic}}
Since $f(x) \sim \text{Normal}(\mu_n(x), \sigma_n^2(x))$, the probability density of $f(x)$ is
  $p\left(f(x)=z\right) = \frac{1}{\sqrt{2\pi}} \exp\left((z - \mu_n(x))^2/2 \sigma_n(x)^2\right)$.
We use this to calculate $\EI(x)$:
\begin{equation*}
\begin{split}
  \EI(x) &= E_n[ (f(x)-f^*_n)^+] \\
  &= \int_{f^*_n}^{\infty} (z-f^*_n) \frac{1}{\sqrt{2\pi}\sigma_n(x)} e^{\frac{-(z - \mu_n(x))^2}{2 \sigma_n^2(x)}} dz \\
  &= \int_{f^*_n}^{\infty} z \frac{1}{\sqrt{2\pi}\sigma_n(x)} e^{\frac{-(z - \mu_n(x))^2}{2 \sigma_n^2(x)}} dz - f_n^*\left(1-\Phi\left(\frac{f_n^*-\mu_n(x)}{\sigma_n(x)}\right)\right) \\
  &= \int_{f^*_n}^{\infty} \left( \mu_n(x) + (z-\mu_n(x)) \right) \frac{1}{\sqrt{2\pi}\sigma_n(x)} e^{\frac{-(z - \mu_n(x))^2}{2 \sigma_n^2(x)}} dz - f_n^*\left(1-\Phi\left(\frac{f_n^*-\mu_n(x)}{\sigma_n(x)}\right)\right) \\
  &= \int_{f^*_n}^{\infty} \left( z-\mu_n(x) \right) \frac{1}{\sqrt{2\pi}\sigma_n(x)} e^{\frac{-(z - \mu_n(x))^2}{2 \sigma_n^2(x)}} dz + (\mu_n(x)-f_n^*)\left(1-\Phi\left(\frac{f_n^*-\mu_n(x)}{\sigma_n(x)}\right)\right) \\
  &= \sigma_n(x)\frac{1}{\sqrt{2\pi}} e^{\frac{-(f^*_n - \mu_n(x))^2}{2 \sigma_n(x)^2}} + (\mu_n(x)-f_n^*)\left(1-\Phi\left(\frac{f_n^*-\mu_n(x)}{\sigma_n(x)}\right)\right) \\
  &= (\mu_n(x)-f_n^*)\left(1-\Phi\left(\frac{f_n^*-\mu_n(x)}{\sigma_n(x)}\right)\right) + \sigma_n(x) \varphi\left(\frac{f_n^*-\mu_n(x)}{\sigma_n(x)}\right) \\
  &= (\mu_n(x)-f_n^*)\Phi\left(\frac{\mu_n(x) - f_n^*}{\sigma_n(x)}\right) + \sigma_n(x) \varphi\left(\frac{\mu_n(x) - f_n^*}{\sigma_n(x)}\right).
\end{split}
\end{equation*}

\subsection{Calculation of the KG factor} \label{sec:KGcalculation}
The KG factor \eqref{eq:KG} is calculated by first considering how the quantity $\mu^*_{n+1} - \mu^*_n$ depends on the information that we have at time $n$, and the additional datapoint that we will obtain, $y_{n+1}$.

First observe that $\mu^*_{n+1} - \mu^*_n$ is a deterministic function of the vector $[\mu_{n+1}(x) : x\in A_{n+1}]$ and other quantities that are known at time $n$.  Then, by applying the analysis in Section~\ref{sec:GP-noise}, but letting the posterior given $x_{1:n},y_{1:n}$ play the role of the prior, we obtain the following version of \eqref{eq:GP-noisy-posterior-mean}, which applies to any given $x$,
\begin{equation}
\mu_{n+1}(x) = \mu_n(x) + \frac{\Sigma_n(x,x_{n+1})}{\Sigma_n(x_{n+1},x_{n+1}) + \lambda^2}\left(y_{n+1} - \mu_n(x_{n+1})\right).
  \label{eq:KG_mu}
\end{equation}
In this expression, $\mu_n(\cdot)$ and $\Sigma_n(\cdot,\cdot)$ are given by \eqref{eq:GP-noisy-posterior-mean-multiple} and \eqref{eq:GP-noisy-posterior-covariance-multiple}.

We see from this expression that $\mu_{n+1}(x)$ is a linear function of $y_{n+1}$, with an intercept and a slope that can be computed based on what we know after the $n$th measurement.

We will calculate the distribution of $y_{n+1}$, given what we have observed at time $n$.  First, $f(x_{n+1}) | x_{1:n},y_{1:n} \sim \text{Normal}\left(\mu_n(x_{n+1}), \Sigma_n(x_{n+1}, x_{n+1})\right)$. Since $y_{n+1} = f(x_{n+1}) + \epsilon_{n+1}$, where $\epsilon_{n+1}$ is independent with distribution $\epsilon_{n+1} \sim \text{Normal}(0, \lambda^2)$, we have 
\begin{equation*}
  y_{n+1} | x_{1:n}, y_{1:n} \sim \text{Normal}\left(\mu_n(x_{n+1}), \Sigma_n(x_{n+1}, x_{n+1}) + \lambda^2 \right).
\end{equation*}
Plugging the distribution of $y_{n+1}$ into \eqref{eq:KG_mu} and doing some algebra, we have
\begin{equation*}
  \mu_{n+1}(x) | x_{1:n}, y_{1:n} \sim \text{Normal} \left( \mu_n(x), \widetilde{\sigma}^2(x, x_{n+1}) \right),
\end{equation*}
where $\widetilde{\sigma}(x, x_{n+1}) = \frac{\Sigma_n(x, x_{n+1})}{\sqrt{\Sigma_n(x_{n+1}, x_{n+1}) + \lambda^2}}$. Moreover, we can write $\mu_{n+1}(x)$ as
\begin{equation*}
  \mu_{n+1}(x) = \mu_n(x) + \widetilde{\sigma}(x, x_{n+1}) Z,
\end{equation*}
where $Z=(y_{n+1}-\mu_n(x_{n+1}))/ \sqrt{\Sigma_n(x_{n+1}, x_{n+1}) + \lambda^2}$ is a standard normal random variable, given $x_{1:n}$ and $y_{1:n}$.

Now \eqref{eq:KG} becomes
\begin{equation*}
\begin{split}
  \KG_n(x) &= E_n \left[ \max_{x'\in A_{n+1}} \mu_n(x') + \widetilde{\sigma}(x', x_{n+1}) Z \mid x_{n+1}=x \right] - \mu^*_n.
\end{split}
\end{equation*}
Thus, computing the KG factor comes down to being able to compute the expectation of the maximum of a collection of linear functions of a scalar normal random variable.
Algorithm~2 of \cite{FrazierPowellDayanik2009}, with software provided as part of the matlabKG library \cite{FrSrcWebsite}, computes the quantity
\begin{equation*}
  h(a,b) = \mathbb{E}\left[\max_{i = 1, \ldots, |a|} (a_i + b_i Z)\right] - \max_{i = 1, \ldots, |a|} a_i
\end{equation*}
for arbitrary equal-length vectors $a$ and $b$.
Using this ability, 
and letting $\mu_n(A_{n+1})$ be the vector $[\mu_n(x'): x'\in A_{n+1}]$ and $\widetilde{\sigma}(A_{n+1},x)$ be the vector $[\widetilde{\sigma}(x', x): x'\in A_{n+1}]$, 
we can write the KG factor as
\begin{equation*}
  \KG_n(x) = h(\mu_n(A_{n+1}),\widetilde{\sigma}(A_{n+1},x))
  + \left[ \max (\mu_n(A_{n+1})) - \mu^*_n\right].
\end{equation*}
If $A_{n+1}=A_n$, as it is in the versions of the knowledge-gradient method proposed in \cite{FrazierPowellDayanik2009,ScottFrazierPowell2011}, then the last term $\max (\mu_n(A_{n+1})) - \mu^*_n$ is equal to $0$ and vanishes.




\bibliographystyle{plain}
\bibliography{pfrazier.bib,additional.bib}

\end{document}